# Precision on Demand: Escalating Weight Bits at Uncertain Decoding Steps in a Small Local Model

*A simulated-quantization study with replay controls on Qwen3-0.6B*

Mohammed Nayeem Khan
mohammed.nayeem.khan@intel.com
mnknayeem@gmail.com

Logesh L
logesh.l@intel.com

Aliyas Ahmad Misri
aliyas.ahmad.misri@intel.com

Intel Corporation

**Abstract**

Local decoding usually fixes one weight precision for the whole generation. We study a different allocation: store 8-bit weights, decode with their top 4 bits, and redo a step at 8 bits only when the low-precision output distribution looks uncertain (entropy above a threshold or a small top-two margin). Quantization is simulated, the model is Qwen3-0.6B, and the benchmark is GSM8K. Cost is counted in weight bytes read per token. On 36724 replayed steps the entropy of the low pass ranks argmax flips with AUC 0.93. At a 15% escalation budget it catches 78% of flips and the margin gate 94%, against the 15% expected for a random gate of equal rate (a computed expectation, not a run). In 150 live problems all-high scores 64.0% and all-low 55.3% (McNemar p=0.053). An entropy gate with a 16% escalation rate scores 62.7% at 0.87 of the all-high weight bytes. Its paired difference is -1.3 (-8.0 to +5.3) points to all-high and +7.3 (+0.0 to +14.7) points to all-low (p=0.080), so the live data do not separate the gate from either end. Verified speculation reaches zero KL at similar bytes in the same cost model. A live random control, a thinking-mode phase rule, a second model and live KL were not run.

**Key findings (recomputed from the frozen result files at build time)**

1. Replay on 36724 steps: low-pass entropy ranks argmax flips with AUC 0.928 (0.924 to 0.932); the margin gives 0.959.
2. At a 15% escalation budget the entropy gate catches 78% of flips and 59% of KL mass, the margin gate 94% and 55%; random catches 15% (expected value, not simulated); the oracle 84% and 85%.
3. Live, n=150: all-high 64.0%, all-low 55.3% (McNemar p=0.053), entropy gate at 16% escalation 62.7% at 0.874 of the all-high weight bytes. The gate's difference to all-high is within noise; its difference to all-low has an interval that starts at +0.0 points (p=0.080).
4. Verified speculation (k=4) needs 0.845 of the all-high bytes per token with zero KL in the byte model; the entropy gate at 15% needs 0.867 with KL 0.0084 nats.
5. The bf16 output head is read in every pass and makes a low pass cost 0.72 of a high pass for a 0.6B model, where 4-bit blocks alone would give 0.53.
6. Not run: live random control, thinking-mode phase gating, live KL, 1.7B, second benchmark. The paper states these gaps wherever a claim depends on them.

# 1 Introduction

A language model that runs on one machine spends most of its time reading weights from memory. For small models decoded one token at a time, the number of bytes read per token is a good first-order measure of cost, and unlike wall-clock time it does not change with the load on the machine. Low-bit weights cut those bytes at some cost in accuracy, and the usual choice is a single precision for the whole model and the whole generation.

This paper tests another way to spend the same bytes. The model keeps one stored copy of the weights at 8 bits and decodes every step with only the top 4 bits of each code. When a low-bit step looks uncertain, the step is redone with all 8 bits and that answer is used. The rule for "uncertain" is cheap because it reads only the output distribution of the low pass: either the entropy of that distribution or the gap between its two largest logits.

The main question is whether, at a given escalation rate, an uncertainty gate recovers more of the all-high behaviour than escalating at random steps, and what that recovery costs in weight bytes. A second question concerns Qwen3 thinking mode, where a long reasoning span precedes a short answer: a phase gate would decode the span at low precision and the answer at high precision. The thinking-mode run did not take place, so that question stays open here.

## 1.1 Scope, data and status

Everything reported comes from one model (Qwen3-0.6B), one benchmark (GSM8K) and one simulated quantizer. A live run decodes 150 test problems under ten configurations and scores the answers. A replay analysis takes the all-high trajectories of 150 test problems (36724 generated steps) and asks what each gate would have escalated if it had been attached to them.

The live run covers 150 problems and no more, so the sample is small and the intervals are wide. Table 1 lists every stage of the plan and whether it ran, and the claims in the paper are limited to the rows marked done.

**Table 1.** Plan versus what ran. Every statement in the paper is limited to the stages that ran.

| Stage | Status | Size |
|---|---|---|
| Calibration pass (all-high, 32 training items, thresholds) | done | 7325 generated tokens |
| Live decoding: all-high, all-low, 4 entropy gates, 4 margin gates | done | 150 problems finished in every configuration; 150 with an all-high run |
| Random escalation at matched rate, live | not run | replay control only (Section 6) |
| Lagged-entropy gate that avoids the wasted low pass | not run | none |
| Phase gating for Qwen3 thinking mode ('think low, answer high') | not run | design only (Section 9) |
| Live KL to all-high under each gate | not run | replay KL only (Section 6) |
| Second model, second benchmark, larger model | not run | none |

## 1.2 Summary of findings

In replay, the entropy of the low pass ranks flip positions with an area under the ROC curve of 0.928 (95% interval 0.924 to 0.932), and the margin gives 0.959 (0.956 to 0.962). A random gate gives 0.5 by construction. At a 15% escalation budget the entropy gate catches 78% of the argmax flips and 59% of the KL mass, the margin gate 94% and 55%, and random escalation 15% of each.

In the live run the all-high configuration scores 64.0% and the all-low configuration 55.3% on 150 items. The gap is -8.7 (-16.7 to -0.7) points with an interval that excludes zero, with a lower bound close to zero, and the exact McNemar test gives p=0.053, so even the all-low loss is borderline at this sample size. The entropy gate with a 16% realised rate scores 62.7% and reads 0.874 of the all-high bytes per token in the cost model of Section 3. Its paired difference is -1.3 (-8.0 to +5.3) points to all-high and +7.3 (+0.0 to +14.7) points to all-low (p=0.080). Section 10 discusses what such intervals can support.

## 2 Related work

The closest work in mechanism is Entropy Adaptive Decoding [9], which switches from a smaller to a larger model when a rolling entropy of the logits crosses a threshold and accepts that the output diverges from the large model instead of verifying it. For one Llama pair on MATH, its abstract reports 96.7% of the 11B score while using the 11B model for 43% of tokens. The gate here is the same kind of rule applied to a different switch: bit-planes of one stored weight tensor in place of two separate models. Because a bit-nested switch shares storage and the KV cache, the cost model changes accordingly.

Nested precision is already established as a storage idea. Any-Precision LLM [8] overlays models of 3 to n bits in the footprint of one n-bit model, Matryoshka Quantization [10] trains one checkpoint whose int2 and int4 slices live inside int8, and MatGPTQ [11] does the same post-training with budget-aware per-layer bit search. QuEPT [3] builds one model that switches among bit-widths at run time. These papers answer how to build weights that can be sliced and leave open a per-token policy for choosing the slice while decoding. The substrate used here is much simpler than any of them: round-to-nearest codes, with the low version obtained by dropping the bottom 4 bits.

The speculative family uses a low-bit version of the target as the proposer and then verifies. BitNest [2] embeds the low-precision proposer in the target's own weights and reports 95.2% average acceptance and 1.48 to 1.61 times speedup on 7B to 8B models. QuantSpec [12] pairs 4-bit weights with a hierarchical 4-bit KV cache in a self-speculative scheme with over 90% acceptance, and Quasar [13] quantizes the verification pass. Kim et al. [5] quantize a Qwen3.5-4B target and a trained block-diffusion proposer. Zhang et al. [4] show that tree verification can erase the memory advantage of a 4-bit target, and TriSpec [14] lets a light proxy approve easy proposals so that the full target runs only on uncertain tokens.

Uncertainty signals are used elsewhere too. MARS [6] relaxes rejection sampling when the top candidates of the target have a small margin, and entropy-aware speculative decoding [18] penalises tokens where both models are uncertain. CALM [17] exits layers early on confidence, and Faster Cascades [15] implement a deferral rule through speculative execution. Don't Waste Bits [7] picks a KV-cache bit-width per token with a learned controller that uses an entropy-based feature; among the abstracts reviewed it is the closest per-token precision switch, and it acts on the cache where this paper acts on the weights.

Two empirical papers motivate the entropy signal. Lotfi et al. [1] report that the token-level KL between a quantized and a full-precision model is high exactly where next-token entropy is high, and they cut chain-of-thought length with a logit penalty. Liu et al. [16] study quantized reasoning models and find W8A8 or W4A16 near lossless and lower bit-widths risky. The replay analysis in Section 5 gives a measurement in the same direction for a different model, a different quantizer, and a different metric (argmax flips as well as KL).

### 2.1 Differences from earlier work, and where verification wins

In the abstracts reviewed, none combines a per-token uncertainty gate on weight precision, no verification step, a matched random control, and a byte-based cost account. That statement is limited to the searches logged for this paper, which used titles and abstracts only; full texts were not read, so a relevant detail may be missing.

Verification-based methods win on one property: the output is exactly what the target would have produced under greedy decoding, so accuracy and KL to the target are unchanged by construction. A gate offers no such guarantee, and its error is whatever it misses. Replay in Section 6 gives numbers for this: at similar bytes, verified speculation has zero KL and the gate has a positive KL. Verification also wins when proposals are accepted at a high rate, because one high-precision pass then serves several tokens.

A gate can win in two situations. It needs no second pass over a block of proposals, so it never pays for a rejected block. It also suits settings where acceptance is hard to predict or where the cost of a high-precision pass is dominated by something that a proposal block cannot share. Whether either situation holds for a given model and hardware is open, and this paper tests it only through a byte model.

## 3 Method

### 3.1 Simulated nested quantization

Every linear weight of the transformer blocks (query, key, value, output, gate, up and down projections) is quantized once, in memory, with round-to-nearest on an asymmetric 8-bit grid, and groups of 64 consecutive input weights share a scale and an offset. The high version dequantizes all 8 bits. The low version keeps the top 4 bits of each code and reconstructs at the midpoint of the dropped cell. Both come from the same stored byte, so no extra weights are held for the low model. The embedding table and the output head stay in bf16.

This is a simulation: the dequantized weights are bf16 tensors and the matrix products run in bf16. No packed kernel exists, so no speedup is measured. The cost of a configuration is computed from the numbers of low and high passes it makes, multiplied by the bytes each pass would read.

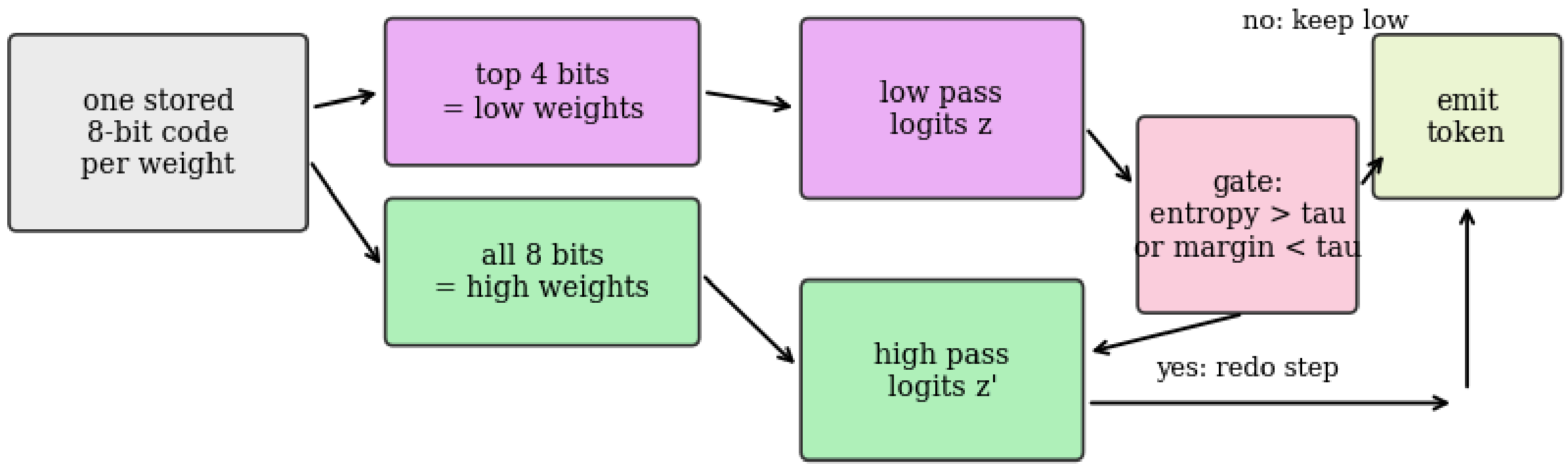

**Figure 1.** Both precisions are read from one stored 8-bit code. The gate looks only at the low-pass distribution; if it fires, the step is redone with the full code.

### 3.2 The gates

Both gates look at the low-pass logits z of the current step. The entropy gate escalates when the entropy H of softmax(z) exceeds a threshold, and the margin gate escalates when the difference between the largest and second-largest logit falls below a threshold. Both thresholds are set on a calibration run of 32 GSM8K training problems decoded with the all-high model, chosen so that the low-pass signal crosses the threshold on 3, 7, 15 and 30 percent of calibration steps. The entropy thresholds are 1.55, 1.16, 0.76, 0.31 nats and the margin thresholds are 0.25, 0.50, 1.25, 2.88, and the names ent0 to ent3 and mar0 to mar3 follow that order.

Escalation is decided after the low pass, so every step pays for a low pass and an escalated step pays for a high pass as well. This waste is part of the cost accounted below. A variant that reuses the previous step's signal would avoid it and was not run.

### 3.3 Cost model and metrics

The cost of one pass is the weight bytes it reads. Block weights cost 8.5 bits per weight in the high version and 4.5 bits in the low version, including 32 bits of scale and offset per group of 64. The bf16 output head is read in every pass. Table 2 lists the byte counts.

**Table 2.** Bytes read per forward pass, from the quantized layout. The head is bf16 in both versions.

| Component | Parameters (M) | Bytes per pass (MB) |
|---|---|---|
| Blocks, high (8 bit + metadata) | 440.4 | 467.9 |
| Blocks, low (4 bit + metadata) | 440.4 | 247.7 |
| Output head (bf16) | 155.6 | 311.2 |
| Whole pass, high | | 779.1 |
| Whole pass, low | | 558.9 |
| Low as share of high | | 0.717 |

Because the head is read in full in every pass, a low pass costs 0.717 of a high pass, while the block ratio alone is 0.529. The head matters for a 0.6B model, where it is about as large as the blocks. Section 8 varies the head precision in the cost model.

The live runs report accuracy with a bootstrap interval over problems, the paired difference to all-high with a bootstrap interval, and an exact McNemar test on the discordant problems. They also report mean generated tokens, the realised escalation rate, and the relative bytes per generated token, with the prompt (prefill) priced at one high pass for every configuration except all-low. KL to the all-high model is reported from replay only.

### 3.4 Controls

The matched random control escalates each step with a probability equal to the realised rate of the gate. Its expected recall of flip positions equals the rate, and its expected KL after escalation is the all-low KL scaled by one minus the rate, so in replay it is an analytic expectation that needs no sampling and has no interval. A live random control was planned and did not run, so no live comparison of the gate with random escalation exists. An oracle control ranks steps by their true KL to the high pass and bounds what any gate based on the low pass could catch.

### 3.5 Phase gating (design)

Qwen3 in thinking mode writes a reasoning span and closes it with a dedicated token before the final answer. A phase gate runs the span at low precision and switches to high precision after that token. The reverse assignment and a combined phase-plus-entropy rule are natural controls, and the driver supports these modes, but none was run on thinking-mode generations, so no result is given.

## 4 Setup

### 4.1 Model, data, decoding

The model is Qwen3-0.6B [20] with its published weights, and the benchmark is GSM8K [19] with problems drawn from the test split by a fixed shuffle. Thresholds come from 32 problems of the training split. The prompt asks for step-by-step reasoning with the final answer in a boxed expression, in the non-thinking chat format, and generation is greedy with a cap of 384 tokens. An answer is correct when the last boxed number after the optional closing think tag equals the gold number. A generation that reaches the cap usually has no boxed answer and scores wrong, so Figure 17 shows how often each configuration hit it.

Generations of different problems share a batch of 12 sequences, each with its own position and its own gating rule, and a finished sequence is replaced by a new problem. Results are written to disk every 20 finished generations, so a stopped run can resume.

### 4.2 Which items are compared

The live run processes problems in order and runs ten configurations on each, so every table of live results uses the 150 problems that finished under all ten configurations and every comparison is paired. The replay analysis uses the all-high trajectories of 150 problems, which are the same problems when 150 equals 150.

## 5 Step-level analysis

The replay analysis feeds the all-high trajectory to the model and runs the low pass at every position, recording the entropy and the margin of the low-pass distribution, the KL from the high to the low distribution, and whether the two argmax tokens agree. A step where they disagree is called a flip. Over 36724 steps from 150 problems the mean KL is 0.0203 nats and 3.79% of steps flip.

Flips are rare and each one can change the rest of a generation, which is why the gates aim at a small fraction of steps. Figure 2 shows the entropy of the low pass separately for steps that flip and steps that do not, with the calibration thresholds marked. Figure 3 shows the margin the same way.

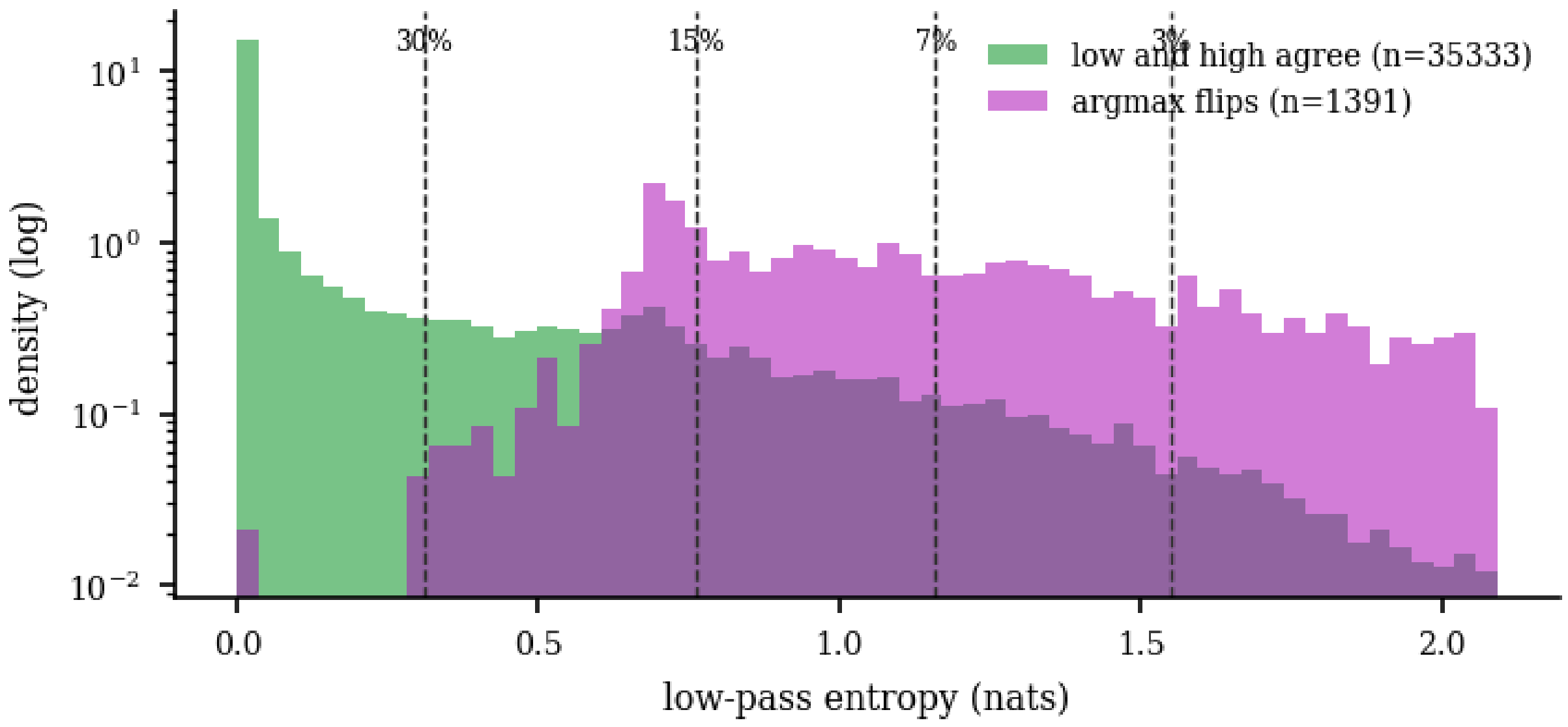

**Figure 2.** Entropy of the low pass, split by whether the argmax flips. Dashed lines are the thresholds that select 3, 7, 15 and 30 percent of calibration steps. Log scale.

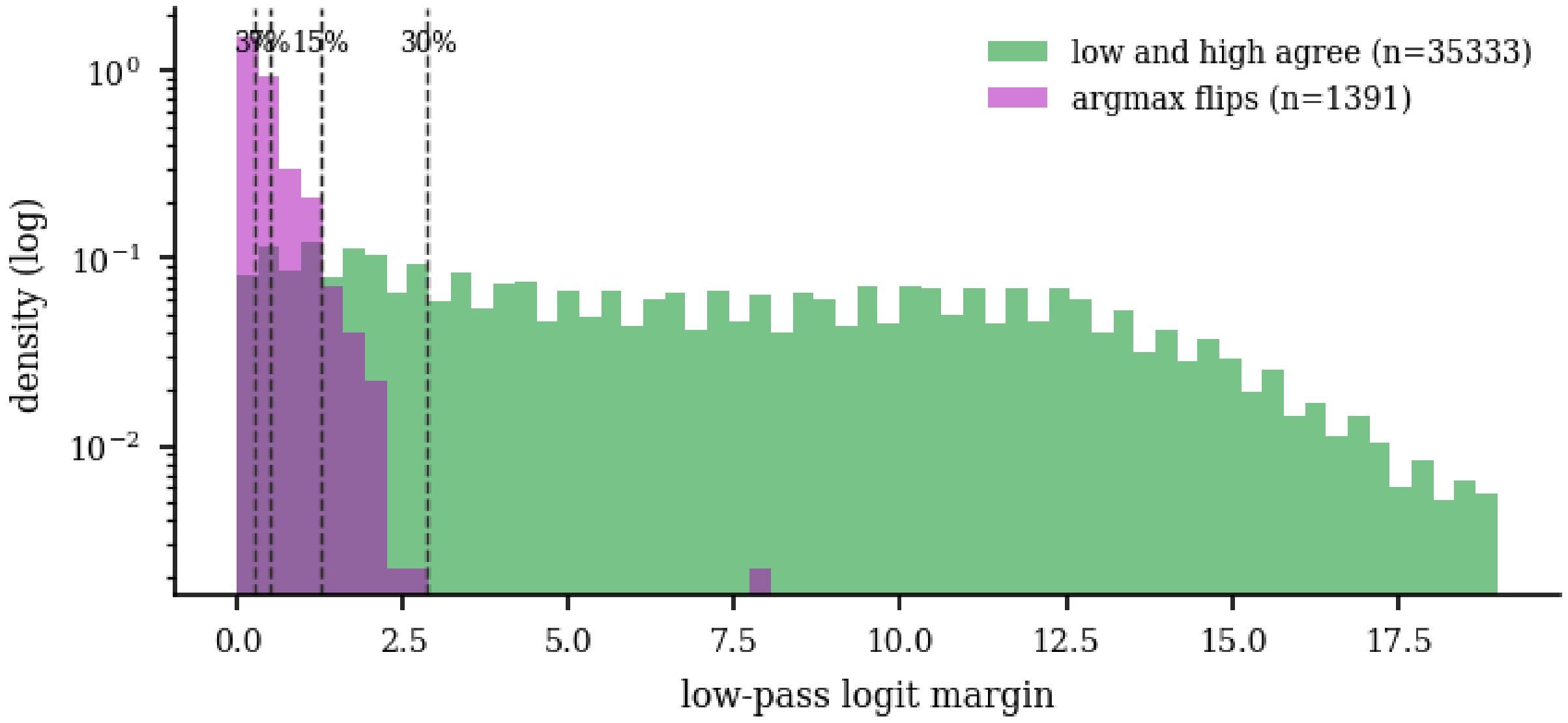

**Figure 3.** Margin of the low pass, split by flip. Dashed lines mark the margin thresholds for the same four rates.

The area under the ROC curve for predicting a flip is 0.928 for entropy and 0.959 for a small margin, with intervals from 100 resamples of problems of 0.924 to 0.932 and 0.956 to 0.962. For predicting the top 10% of steps by KL the values are 0.924 and 0.908. These numbers describe ranking quality over all steps, and the gate uses only the extreme tail.

### 5.1 KL and flips against entropy

Figure 4 shows how the mean KL between high and low and the flip rate change with the entropy of the low pass, and Figure 5 gives the same information by decile. The direction agrees with the link between entropy and quantization divergence reported by Lotfi et al. [1] for other models and quantizers.

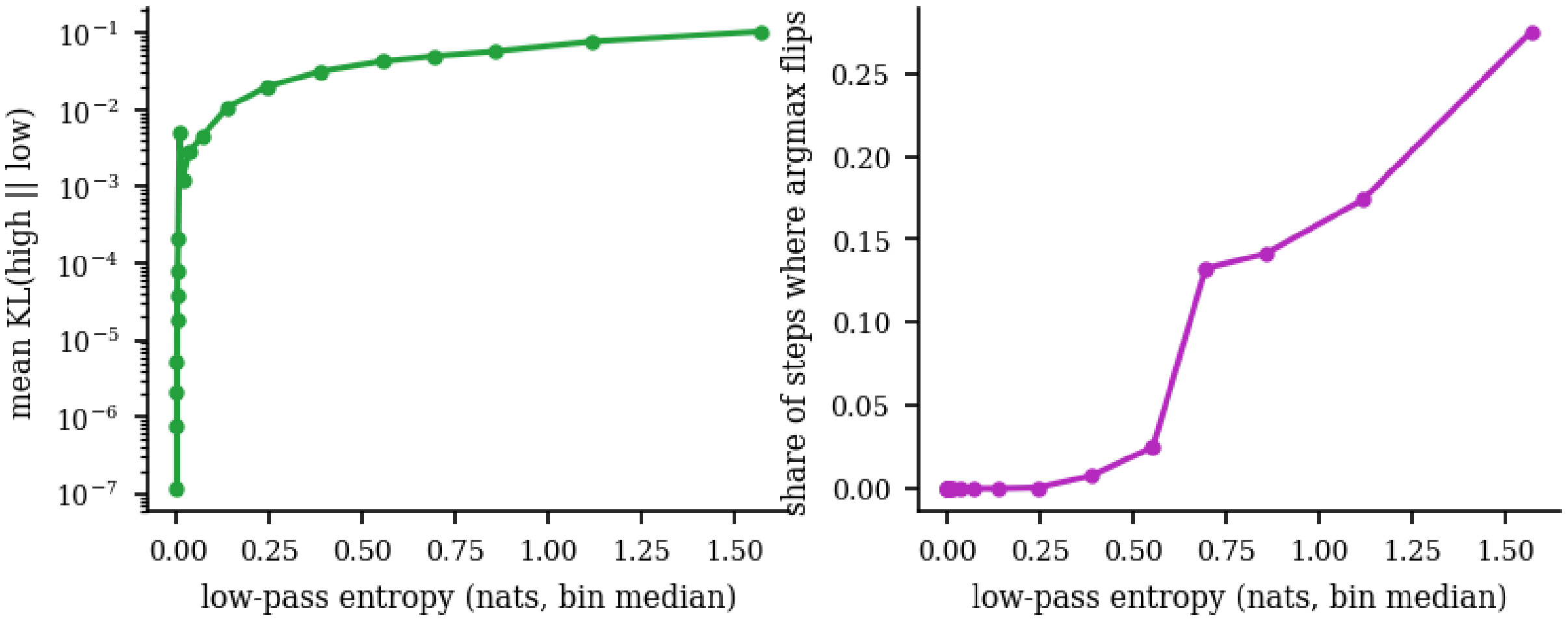

**Figure 4.** Left: mean KL(high || low) in 20 entropy quantile bins. Right: share of steps whose argmax flips in the same bins.

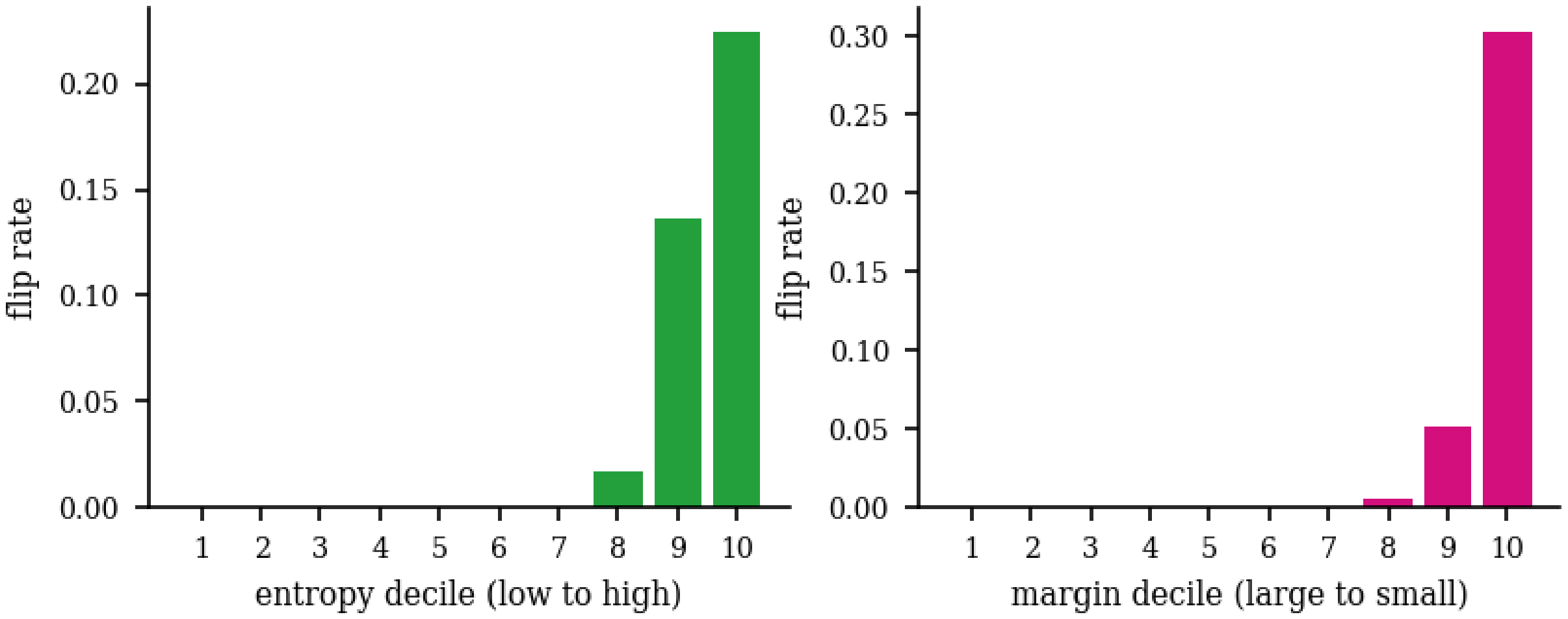

**Figure 5.** Flip rate by decile of the entropy (left) and of the inverted margin (right). Decile 10 holds the most uncertain steps.

Entropy and margin carry overlapping information, and Figure 6 plots the flip rate over both. A step can have a small margin with moderate entropy when two tokens split the mass, and a gate on either signal alone treats such a step differently from a step with a flat distribution.

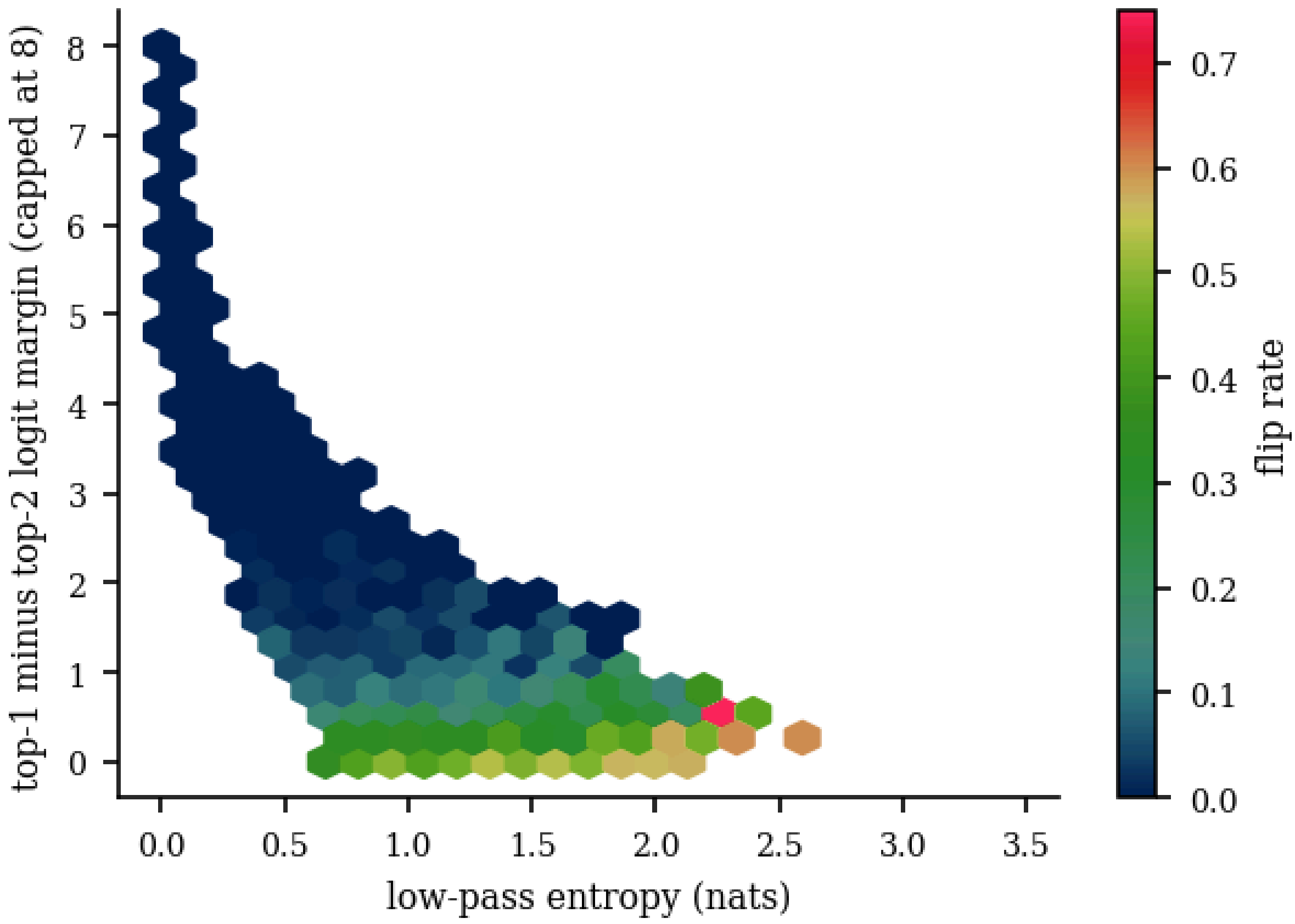

**Figure 6.** Flip rate over the plane of low-pass entropy and margin (hexagonal bins with at least 8 steps).

### 5.2 Which tokens flip

Token types may differ in how often they flip. Figure 7 groups generated tokens into numbers, operators, words, punctuation, newlines and mixed pieces; numbers and operators carry the arithmetic of a GSM8K solution, so an error there changes the answer directly. The right panel compares the share of each class among all steps with its share among the steps that the ent2 gate escalates.

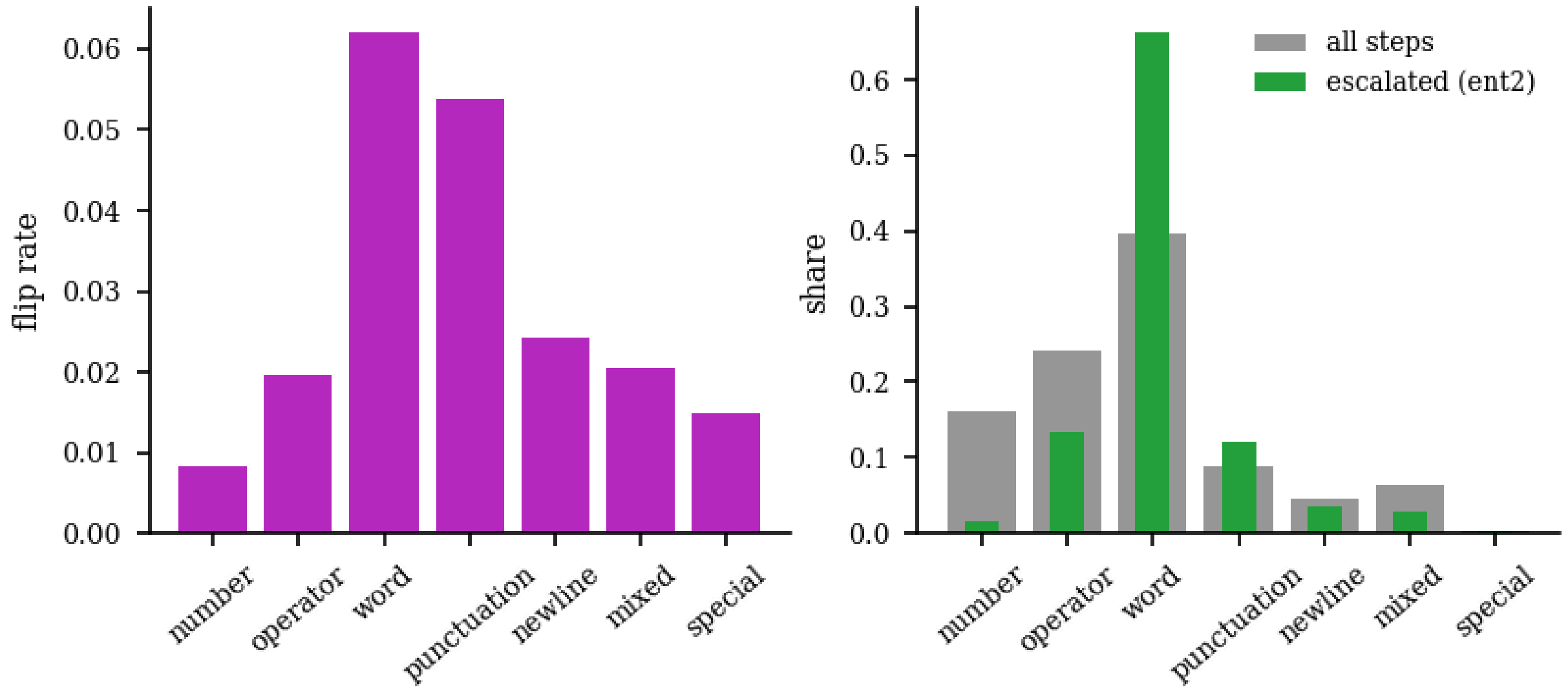

**Figure 7.** Flip rate by token class (left) and share of each class among all steps and among ent2 escalations (right). Classes with fewer than 20 steps are omitted.

### 5.3 Position in the generation

A phase rule assumes that different parts of a generation differ in sensitivity to precision. Figure 8 shows entropy, flip rate and KL against the relative position inside a generation, which gives a weak proxy for that assumption on non-thinking generations.

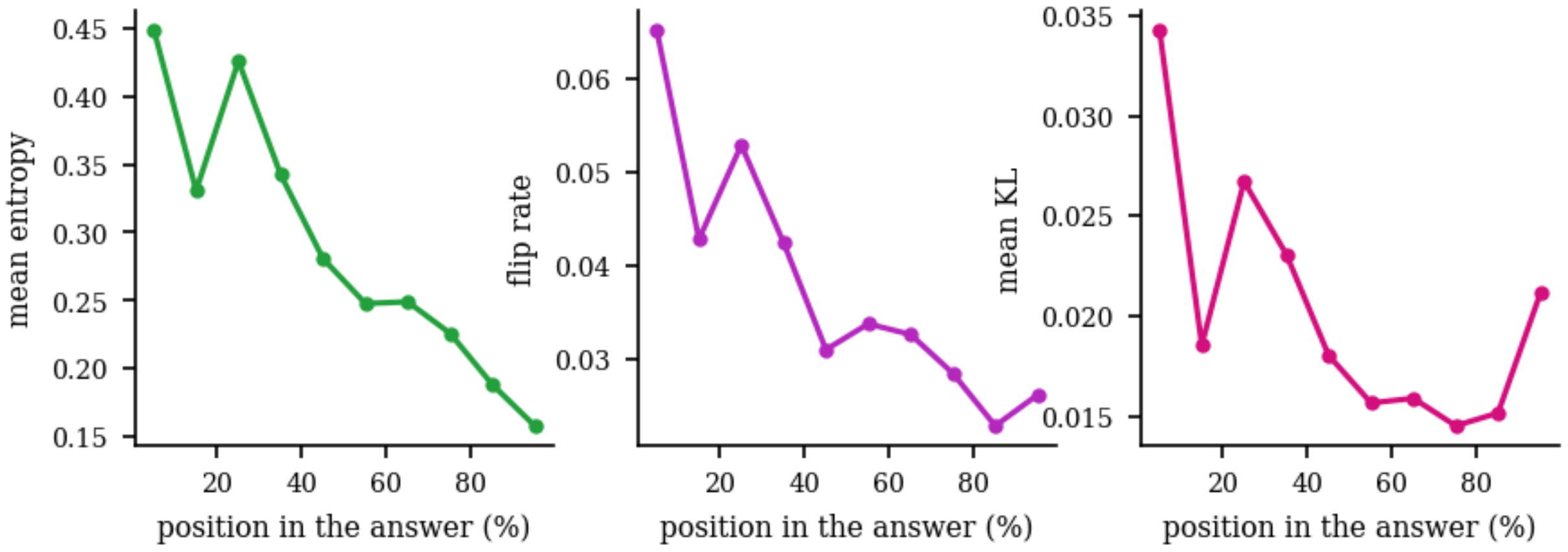

**Figure 8.** Mean entropy, flip rate and mean KL by tenth of the generation (replay on the all-high trajectories).

### 5.4 Where a free-running low model first departs

A free-running all-low generation follows the all-high one until its first different token. In 150 problems of the 150 with an all-low run, the first departure has a median index of 2 tokens. The median entropy percentile of the departing token, among all steps, is 87% (Figure 9). A departure at a step the entropy gate would have escalated can be repaired in a live run, but a departure at a confident step cannot.

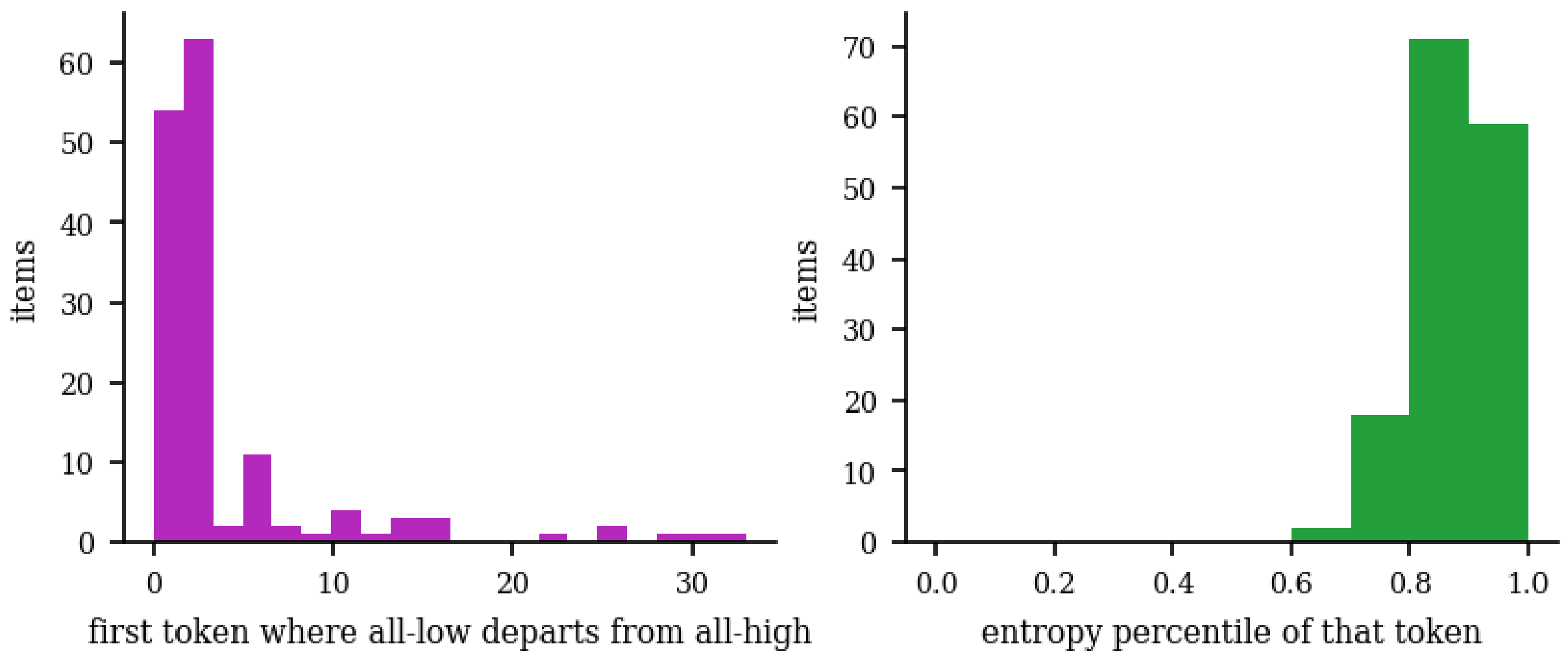

**Figure 9.** Left: index of the first token at which the free-running all-low generation differs from all-high. Right: percentile of the all-high low-pass entropy at that token among all steps.

### 5.5 Worked examples of caught and missed departures

Table 3 gives the share of the 150 first departures that each gate would have escalated. The first departure is the step that starts the divergence of a free-running low model, so a gate that misses it leaves the generation on the low path unless a later step is caught. Table 4 shows four of them, two at steps that the ent2 gate would escalate and two at steps it would miss. The first-departure index has a median of 2 over all items, so the examples are later than typical: they are restricted to index 6 or later so that the context is readable, and within that set the two nearest to the median index are taken, not hand-picked.

**Table 3.** Share of first departures (all-low versus all-high) that fall on a step each calibrated gate escalates. The all-high low-pass signals at the departure token are used.

| Gate | Replay rate over all steps | Share of first departures escalated | Ratio |
|---|---|---|---|
| ent0 | 2.6% | 7% | 2.8x |
| mar0 | 2.6% | 18% | 6.8x |
| ent1 | 6.9% | 26% | 3.8x |
| mar1 | 6.1% | 42% | 6.9x |
| ent2 | 14.9% | 58% | 3.9x |
| mar2 | 14.7% | 85% | 5.8x |
| ent3 | 30.1% | 99% | 3.3x |
| mar3 | 29.3% | 99% | 3.4x |

**Table 4.** First departures of a free-running all-low generation. Context is the last eight all-high tokens before the step; line breaks are shown as NL. Entropy in nats and margin in logits are from the low pass on the all-high prefix.

| Problem | Index | Context | All-high token | All-low token | Entropy | Margin | Gates that fire |
|---|---|---|---|---|---|---|---|
| problem 18 | 6 | ...We are told that a candle | ** | melts | 0.94 | 0.50 | ent2, mar2, ent3, mar3 |
| problem 28 | 6 | ...We are given: NL NL - Billy | helps | can | 0.83 | 0.75 | ent2, mar2, ent3, mar3 |
| problem 84 | 6 | ...We are given: NL NL - Each | goat | ** | 0.69 | 0.38 | mar1, mar2, ent3, mar3 |
| problem 66 | 9 | ...'s break down the problem step by step | : NL NL | . NL NL | 0.59 | 1.00 | mar2, ent3, mar3 |

## 6 Controls in replay

Replay works as a controlled experiment here: each gate is attached to the all-high trajectory, and what it would have escalated is compared with random and oracle gates of equal rate. An escalated step is assumed to match the high distribution exactly, so the result is a first-order estimate of KL to all-high that ignores how a flip changes later context in a live run, which Section 7 measures through accuracy.

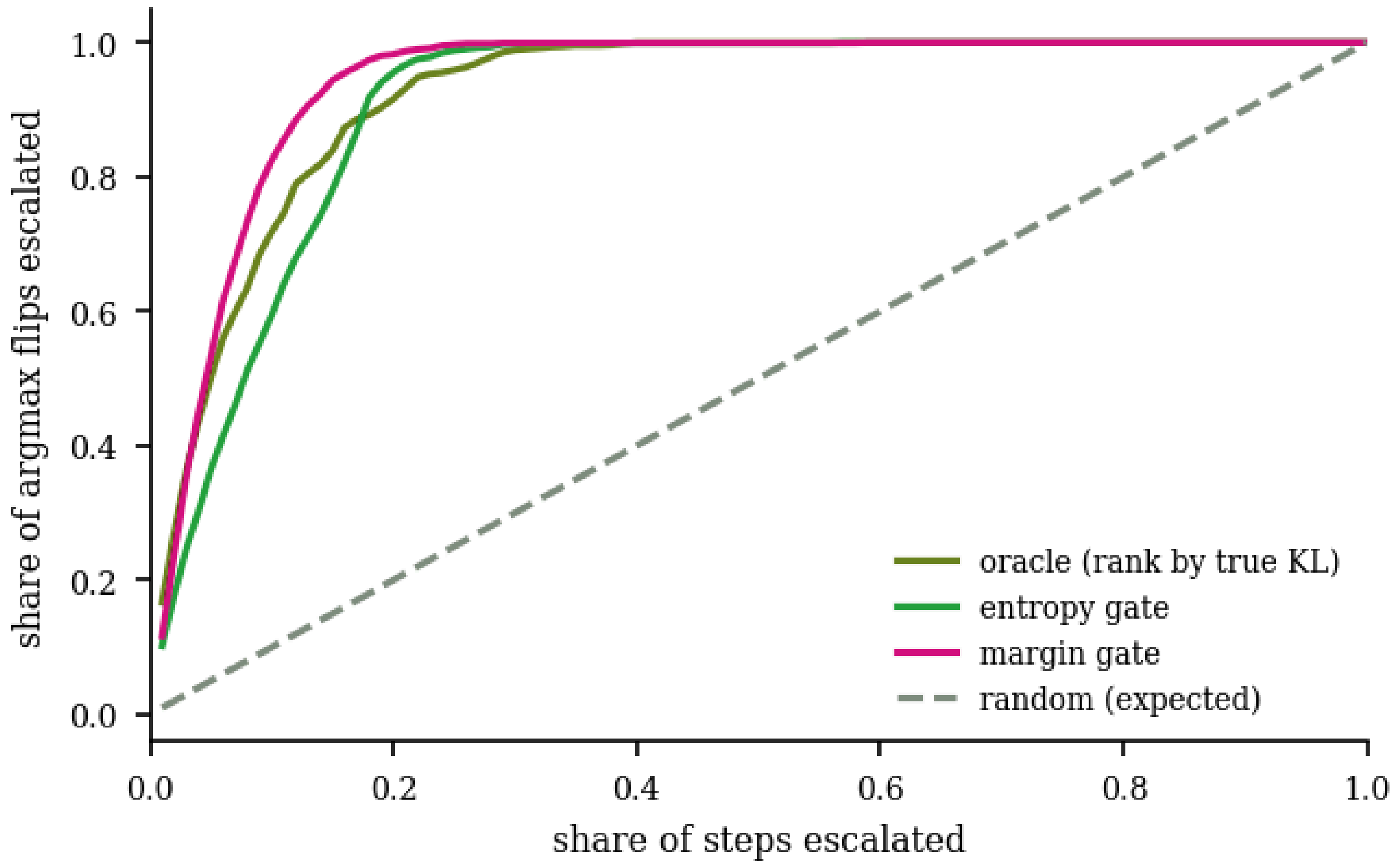

**Figure 10.** Share of flip positions that a gate escalates against the share of steps escalated, for entropy and margin ordering, the random expectation and the oracle ordering.

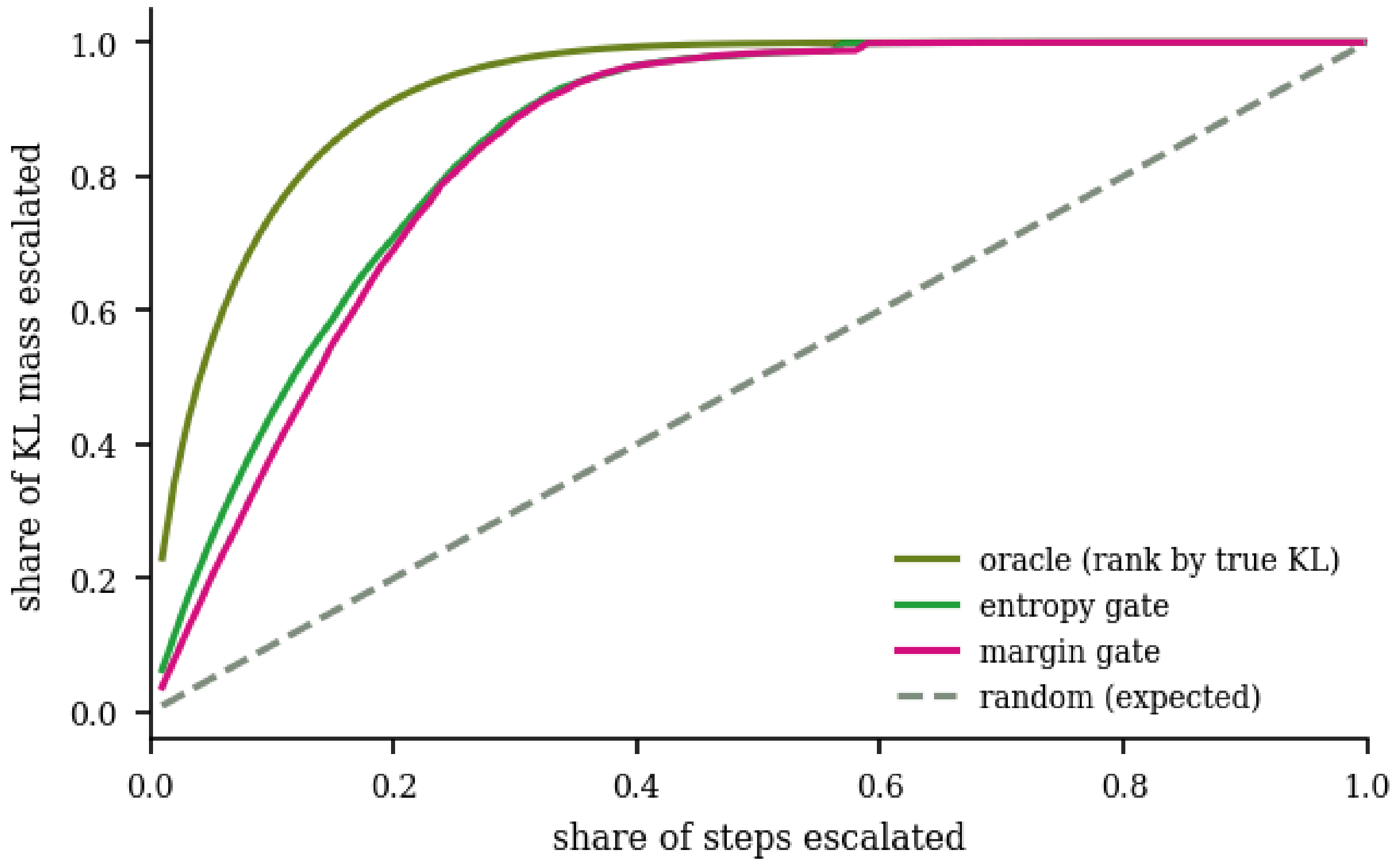

**Figure 11.** The same for the share of KL mass. The oracle ranks steps by their true KL.

Figures 10 and 11 show both gates against the random diagonal and the oracle. At a 15% budget the entropy ordering reaches 78% of flips and 59% of KL mass, the margin ordering 94% and 55%, and the oracle 84% and 85%, while the random control reaches 15% of each. These values are for the pooled replay set, and the interval columns of Table 5 show their spread over problems.

The two orderings trade places depending on the target. For the calibrated 15% gates (ent2 and mar2, with replay rates 14.9% and 14.7%) the margin gate escalates 94% of flips against 78% for the entropy gate, while the entropy gate removes 59% of the KL against 54% for the margin gate. For the 7% gates (ent1 and mar1, rates 6.9% and 6.1%) the figures are 63% against 45% for flips and 33% against 24% for KL. The 55% in the budget curves for the margin ordering differs from the 54% of the calibrated gate only because the calibrated gate escalates 14.7% of steps and not exactly 15%. A reader who cares about the argmax token (and so about the text) and a reader who cares about the whole distribution may therefore prefer different gates.

**Table 5.** Replay of the calibrated gates on 36724 held-out all-high steps from 150 problems. Recall is the share of flips escalated (95% interval over problems); lift is recall divided by rate, which is 1.0 for random. KL removed is the share of total KL in escalated steps. KL left is the mean KL per step after gating, in nats. Bytes are relative to all-high including the wasted low pass.

| Gate | Rate | Precision | Recall of flips (%) | Lift | KL removed (%) | KL left | Bytes |
|---|---|---|---|---|---|---|---|
| ent0 | 2.6% | 32.5% | 23 (20 to 26) | 8.6x | 15 (13 to 16) | 0.0173 | 0.744 |
| ent1 | 6.9% | 25.0% | 45 (42 to 48) | 6.6x | 33 (31 to 35) | 0.0136 | 0.786 |
| ent2 | 14.9% | 19.8% | 78 (76 to 80) | 5.2x | 59 (56 to 60) | 0.0084 | 0.867 |
| ent3 | 30.1% | 12.6% | 100 (100 to 100) | 3.3x | 89 (87 to 91) | 0.0022 | 1.018 |
| mar0 | 2.6% | 46.7% | 33 (30 to 35) | 12.3x | 11 (10 to 12) | 0.0182 | 0.744 |
| mar1 | 6.1% | 38.8% | 63 (60 to 65) | 10.2x | 24 (23 to 26) | 0.0154 | 0.779 |
| mar2 | 14.7% | 24.2% | 94 (92 to 95) | 6.4x | 54 (52 to 56) | 0.0094 | 0.864 |
| mar3 | 29.3% | 12.9% | 100 (100 to 100) | 3.4x | 87 (85 to 89) | 0.0026 | 1.010 |

### 6.1 Cost against KL, with verified speculation

Figure 12 places the gates on a plane of bytes per token against KL to all-high, with verified speculation added from the same trajectories. A round proposes k tokens with low passes, verifies them with one high pass, and accepts the longest agreeing prefix plus one token. The replay uses the argmax agreement of the low and high passes as the acceptance test, which is exact for greedy decoding, so its KL is zero by construction. Its bytes assume that one high pass over k+1 positions costs one pass of weight reads, as in a memory-bound decoder.

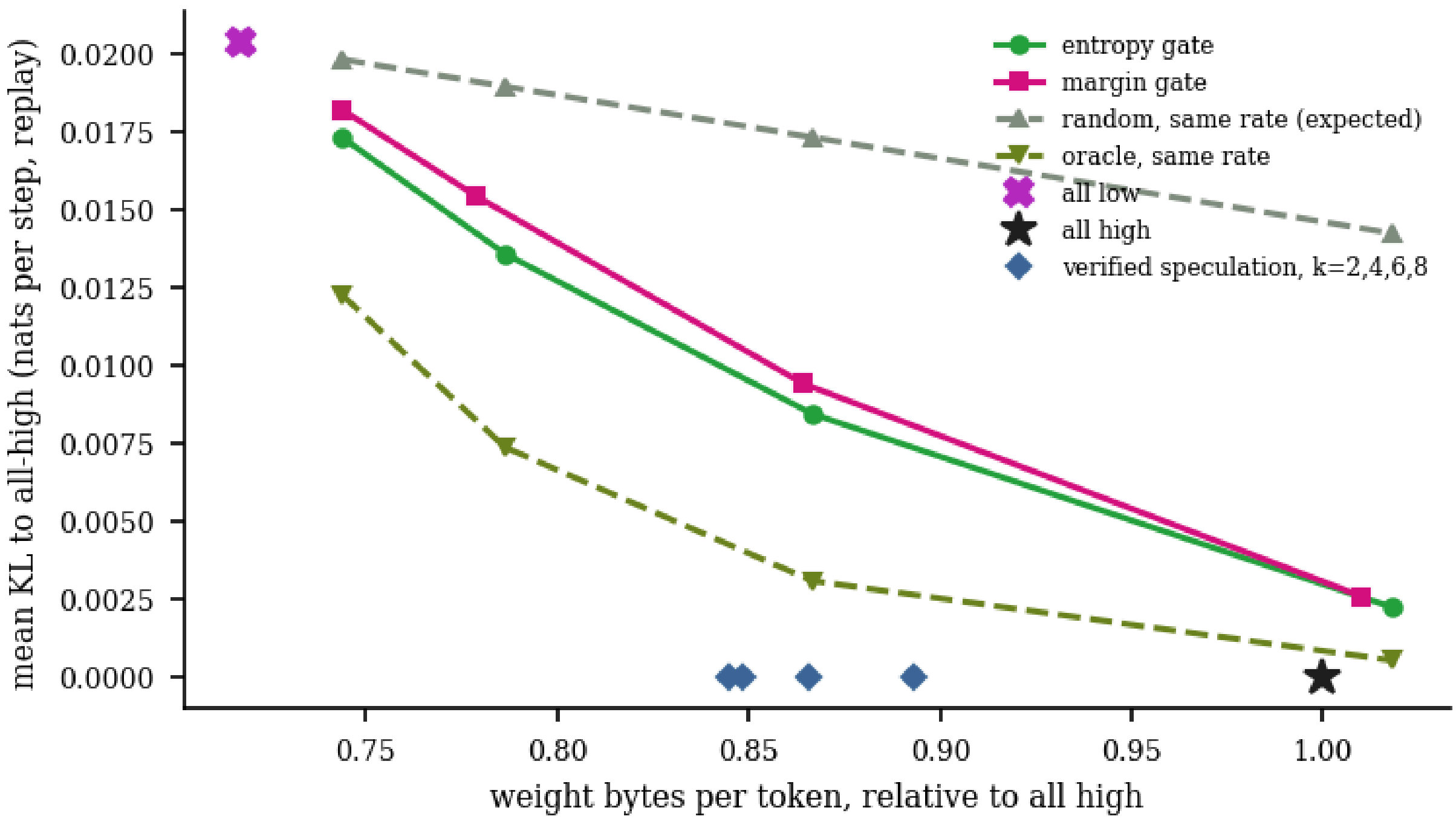

**Figure 12.** Replay frontier. Each gate point is one calibrated threshold. Random and oracle points use the same rates as the entropy gate. Verified speculation sits on the zero-KL line.

In this model of cost, verified speculation with k=4 needs 0.845 of the all-high bytes per token and accepts 4.58 tokens per round. The entropy gate at 15% needs 0.867 and leaves a mean KL of 0.0084 nats. At these byte counts the verified scheme reaches zero KL for similar bytes, so it is the better choice whenever the target can be run for verification and proposals are accepted at the rate seen here. The gate gives up exactness and avoids the machinery of proposal blocks and rollback, which this model prices at zero.

Table 6 gives the replay for several proposal lengths. Bytes per token move with k because every low pass reads the bf16 head.

**Table 6.** Verified speculation replayed on the all-high trajectories (acceptance from argmax agreement). Tokens per round is generated steps divided by rounds, so it sits slightly below mean accepted plus one: the last round of a generation has no bonus token.

| k | Rounds | Mean accepted per round | Tokens per round | Bytes vs all-high |
|---|---|---|---|---|
| 2 | 12793 | 1.88 | 2.87 | 0.848 |
| 4 | 8019 | 3.59 | 4.58 | 0.845 |
| 6 | 5993 | 5.15 | 6.13 | 0.866 |
| 8 | 4868 | 6.57 | 7.54 | 0.893 |

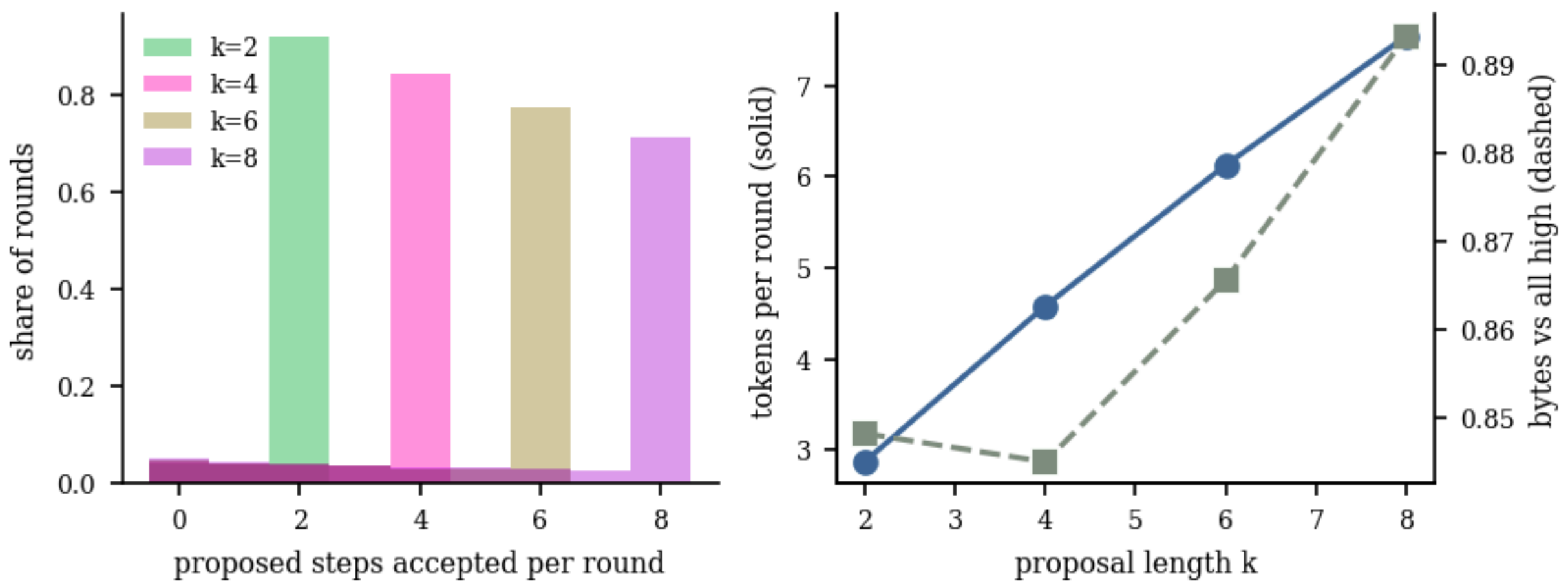

**Figure 13.** Left: distribution of the number of accepted proposals per round. Right: tokens per round and relative bytes against k.

## 7 Live decoding results

This section reports the live decoding results on the 150 problems that finished under all ten configurations. Accuracy intervals are bootstrap intervals over problems, the difference to all-high is paired, and the last column gives the discordant counts (right with all-high and wrong here, then the reverse) and the exact McNemar p-value.

**Table 7.** Live results, Qwen3-0.6B on GSM8K, n=150 problems. Accuracy and paired difference in percent and points with 95% bootstrap intervals. Bytes columns are relative to all-high: blocks only, and blocks plus head.

| Config | Accuracy (%) | Diff to all-high (pts) | Tokens | Escalated | Block bytes | Total bytes | Lost/gained, p |
|---|---|---|---|---|---|---|---|
| high | 64.0 (56.7 to 71.3) | | 245 | 100% | 1.000 | 1.000 | |
| low | 55.3 (47.3 to 63.3) | -8.7 (-16.7 to -0.7) | 258 | 0% | 0.529 | 0.717 | 26/13, p=0.053 |
| ent0 | 62.0 (54.0 to 69.3) | -2.0 (-10.0 to +5.3) | 241 | 3.4% | 0.565 | 0.752 | 19/16, p=0.736 |
| ent1 | 62.0 (54.7 to 70.0) | -2.0 (-8.7 to +4.7) | 246 | 7.3% | 0.604 | 0.791 | 14/11, p=0.690 |
| ent2 | 62.7 (55.3 to 70.7) | -1.3 (-8.0 to +5.3) | 245 | 15.7% | 0.687 | 0.874 | 13/11, p=0.839 |
| ent3 | 65.3 (58.0 to 72.7) | +1.3 (-5.3 to +7.4) | 240 | 31.0% | 0.840 | 1.027 | 10/12, p=0.832 |
| mar0 | 59.3 (52.0 to 67.3) | -4.7 (-12.0 to +2.7) | 240 | 2.9% | 0.560 | 0.747 | 20/13, p=0.296 |
| mar1 | 61.3 (53.3 to 68.7) | -2.7 (-10.0 to +4.7) | 245 | 6.1% | 0.592 | 0.779 | 18/14, p=0.597 |
| mar2 | 66.0 (58.7 to 73.3) | +2.0 (-5.3 to +9.3) | 240 | 15.0% | 0.681 | 0.868 | 12/15, p=0.701 |
| mar3 | 64.7 (57.3 to 72.7) | +0.7 (-5.3 to +6.7) | 240 | 29.6% | 0.827 | 1.014 | 10/11, p=1.000 |

All-high scores 64.0% and all-low 55.3%, and all-low differs from all-high by -8.7 (-16.7 to -0.7) points with an interval that excludes zero. The ent2 gate scores 62.7% at 0.874 of the all-high bytes and the mar2 gate 66.0% at 0.868; their paired differences to all-high are -1.3 (-8.0 to +5.3) and +2.0 (-5.3 to +9.3) points. Against all-low the ent2 gate differs by +7.3 (+0.0 to +14.7) points (McNemar p=0.080), whose lower bound is +0.00 points, and the mar2 gate by +10.7 (+2.7 to +18.7) points (McNemar p=0.017). These are two of eight gate-versus-all-low comparisons and none is corrected for that, so only the mar2 comparison is clear of zero by a margin, and mar2 is the highest-scoring configuration of the ten, which inflates its estimate. With 150 problems, an interval that includes zero says only that the sample cannot separate two configurations, and it leaves open whether they are equal.

The live bytes include one high prefill pass per problem and the replay bytes do not, and the live escalation rates differ from the replay rates (Figure 16), which is why ent2 reads 0.874 here and 0.867 in Table 5. ent3, mar3 read more than the all-high bytes per token, because a gate at a high rate pays for a low pass on every step and a second pass on each escalated one. Escalating a third of all steps with this form of the gate is a loss in the byte model whatever its accuracy, and the lagged variant mentioned in Section 3.2 would reduce that waste but was not run.

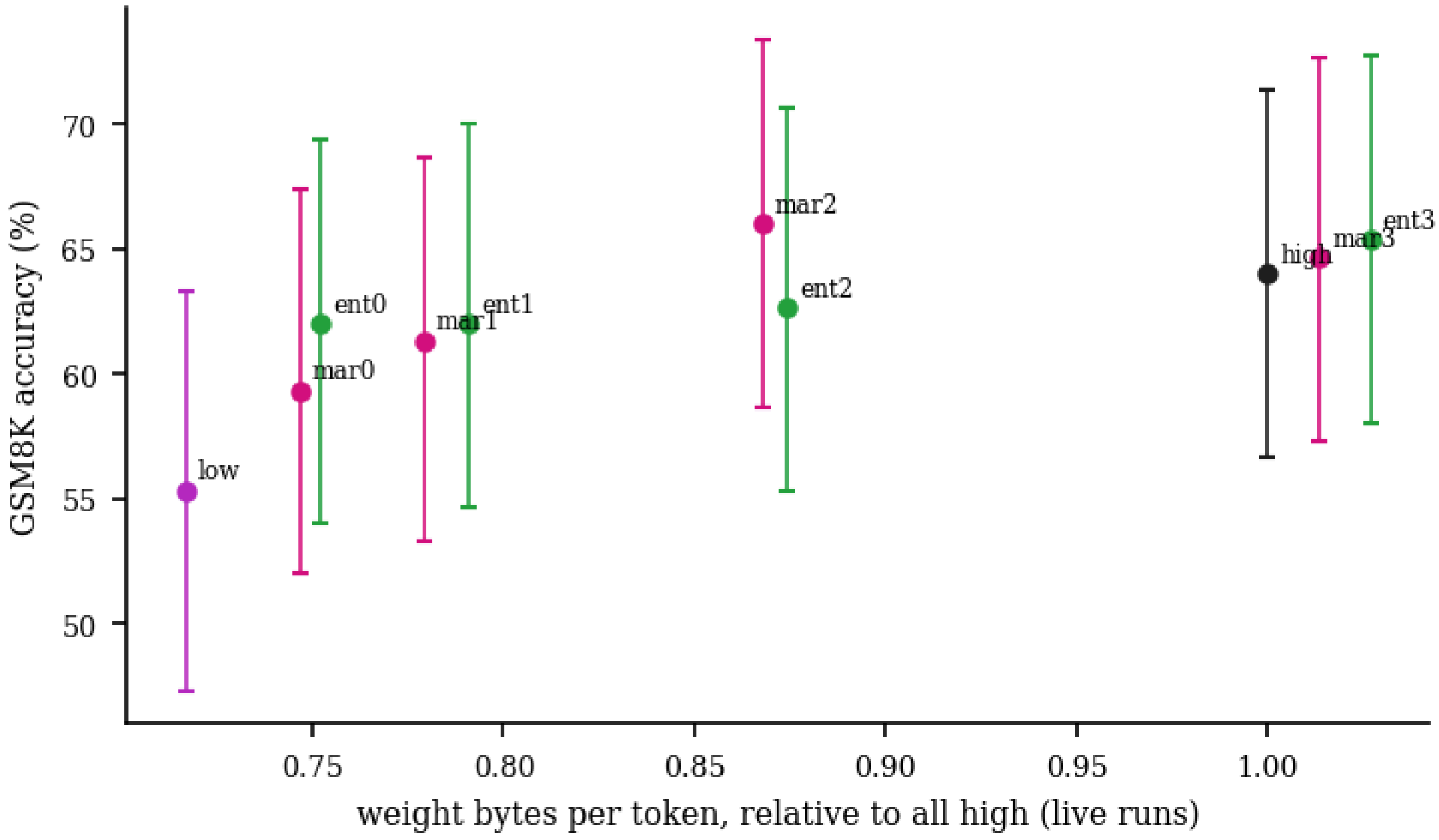

**Figure 14.** Accuracy with 95% bootstrap intervals against relative bytes per token for every live configuration.

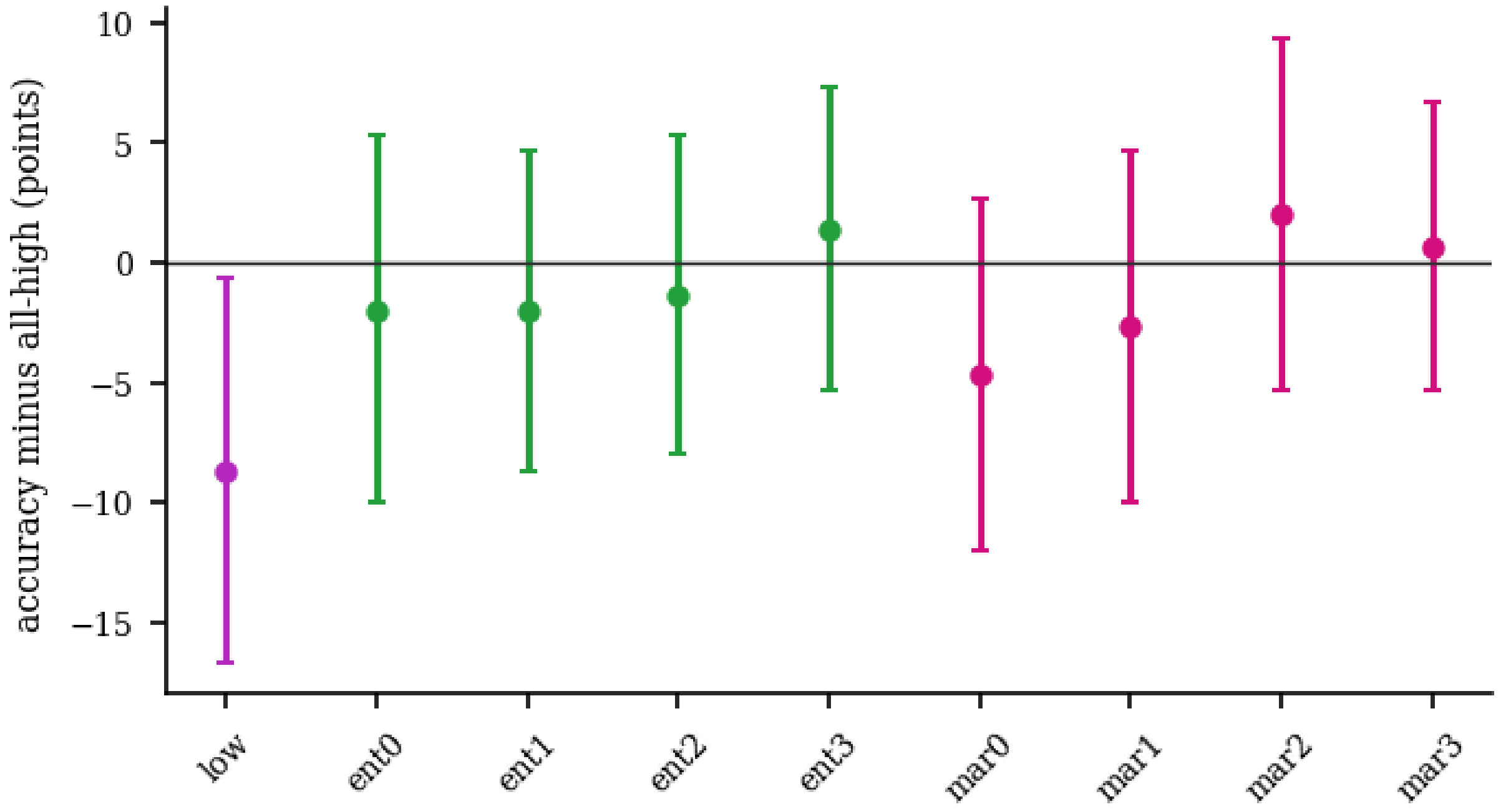

**Figure 15.** Paired difference to all-high with 95% bootstrap intervals. The zero line is all-high.

## 7.1 Escalation rate, length and discordant items

The realised escalation rate in the live run need not equal the rate on the calibration set, because the live trajectories differ and because thresholds were fitted on 32 other problems. Figure 16 compares the targets with the live and replay rates.

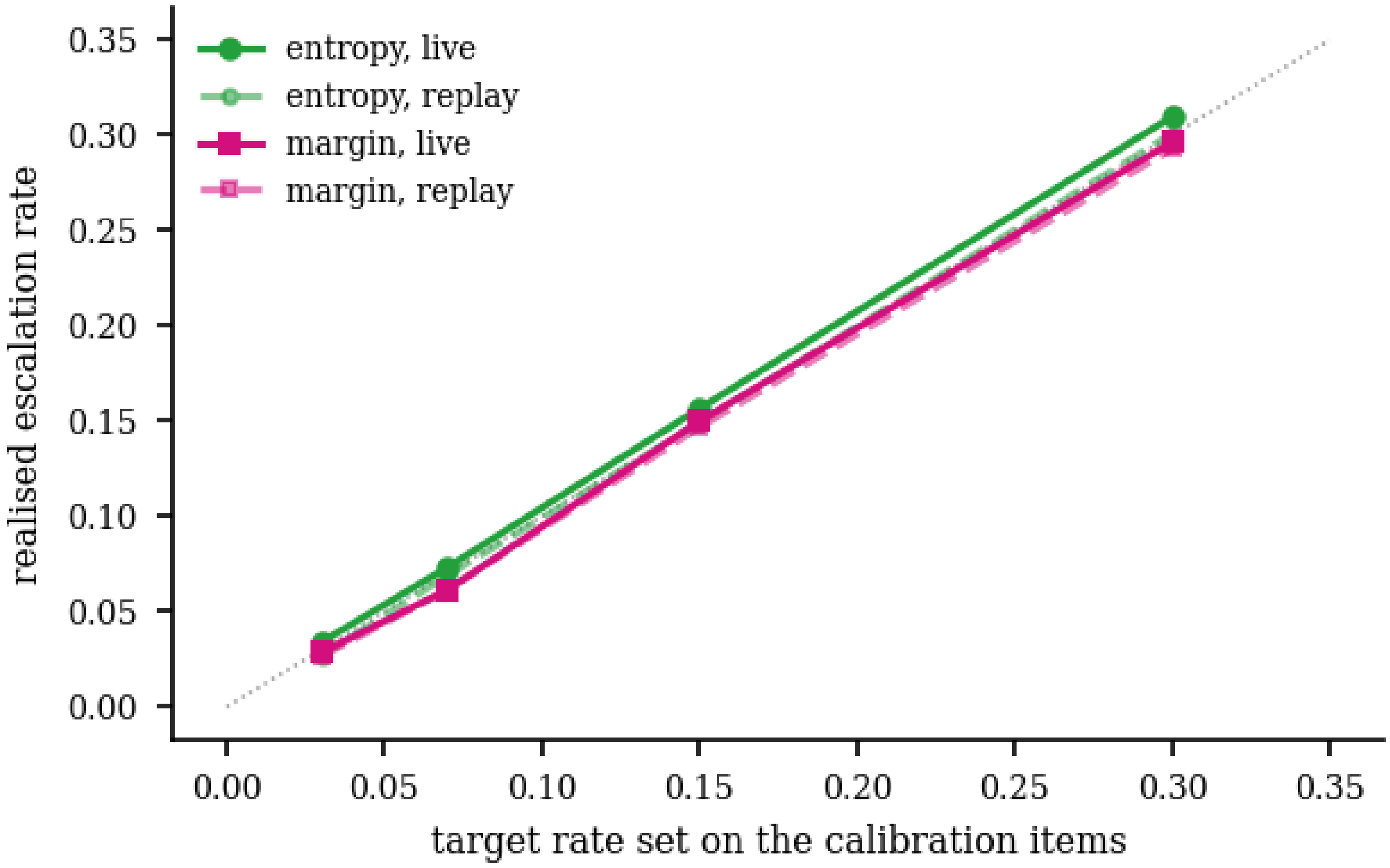

**Figure 16.** Target rate (calibration set) against the realised rate in live runs and in replay on all-high trajectories.

Lower precision also changes how long a solution is and how often it runs into the cap. Figure 17 shows mean generated tokens and the share of capped generations per configuration; token counts are a cost on their own, since each token needs a pass.

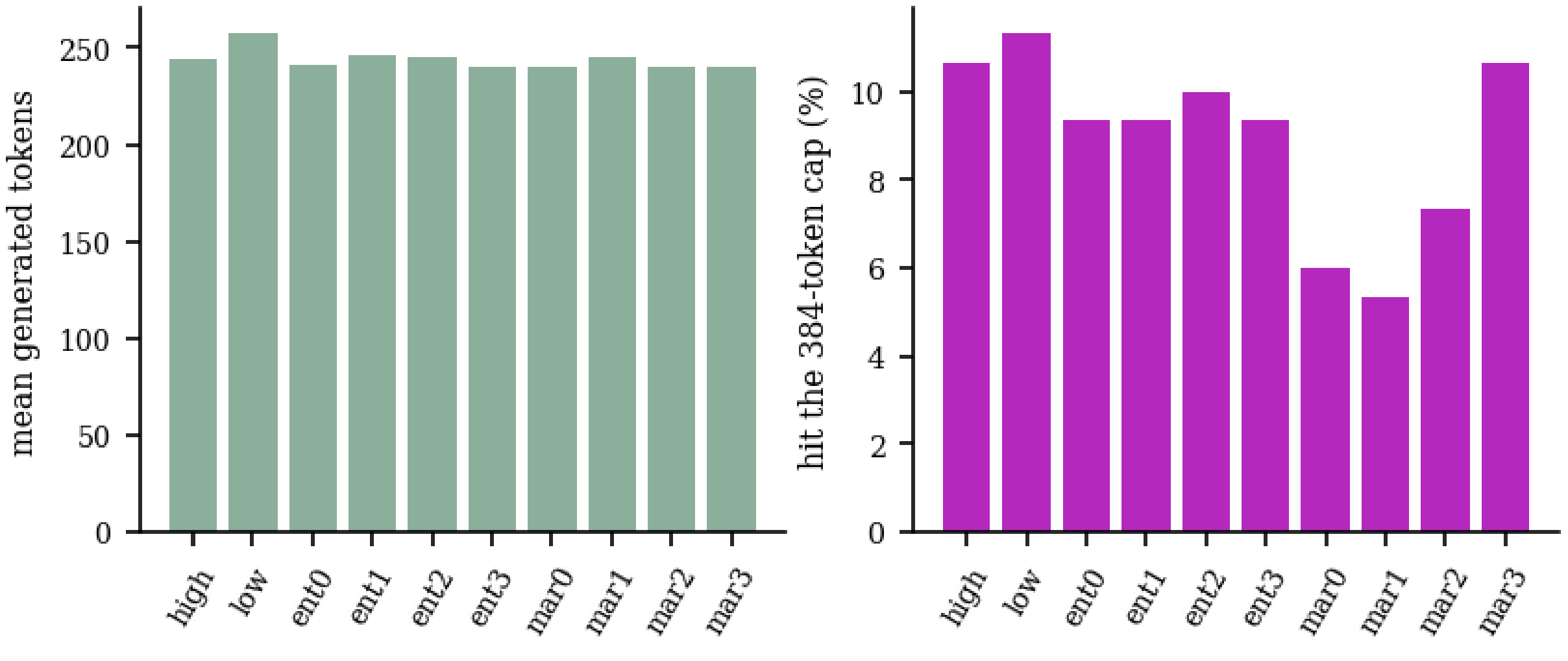

**Figure 17.** Mean generated tokens (left) and share of generations that hit the 384-token cap (right).

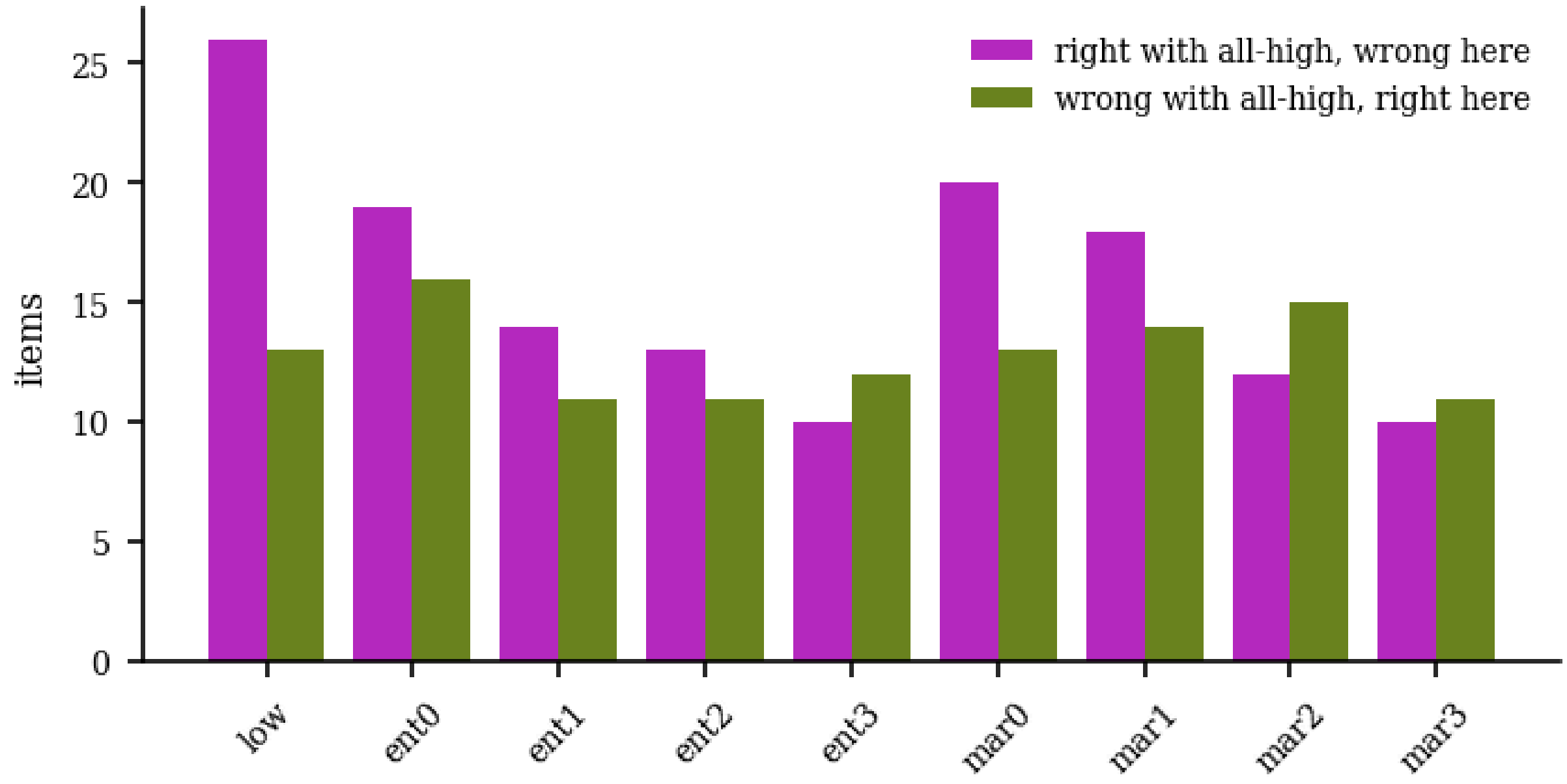

**Figure 18.** Items that flip between right and wrong relative to all-high, per configuration.

Configurations with lower escalation rates share more of their trajectories with all-low, so their discordant items are partly the same problems and the counts in Figure 18 are correlated across configurations. The McNemar tests are run per configuration and are not corrected for multiple comparisons.

## 8 Cost model

The cost axis is the weight bytes of a memory-bound decoder, and three facts of the model matter. The output head is read in every pass and makes up 40% of a high pass and 56% of a low pass at this model size. The gate pays for a low pass on every step before it knows whether to escalate. In verified speculation the proposal passes also read the head.

Figure 19 breaks one generated token of four schemes into block bytes at low and high precision and head bytes. Figure 20 changes the head precision in the model, with all passes using the same head; accuracy under a lower-precision head was not measured, so this is a cost-model what-if.

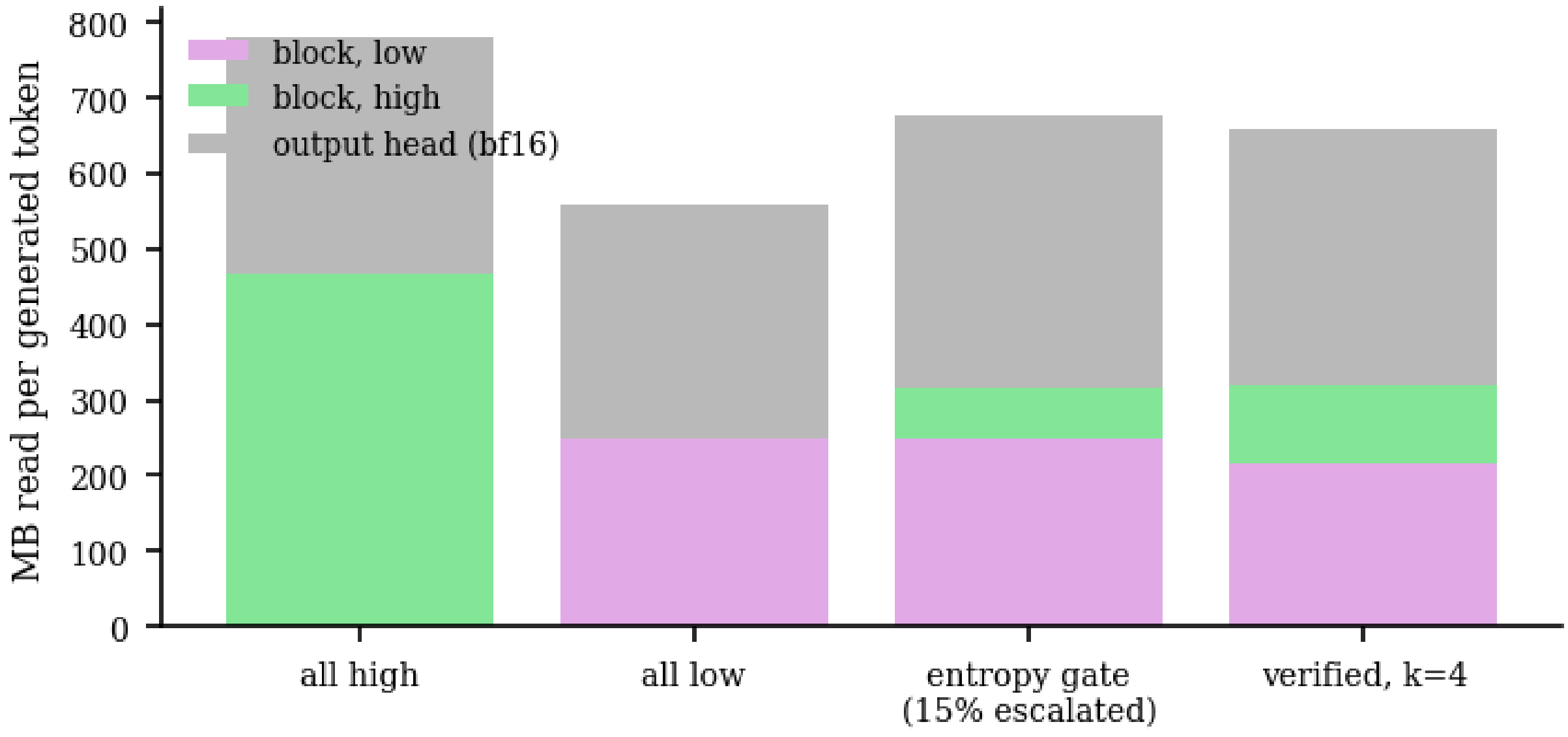

**Figure 19.** Bytes read per generated token, split into low blocks, high blocks and the bf16 head, for four schemes (replay rates).

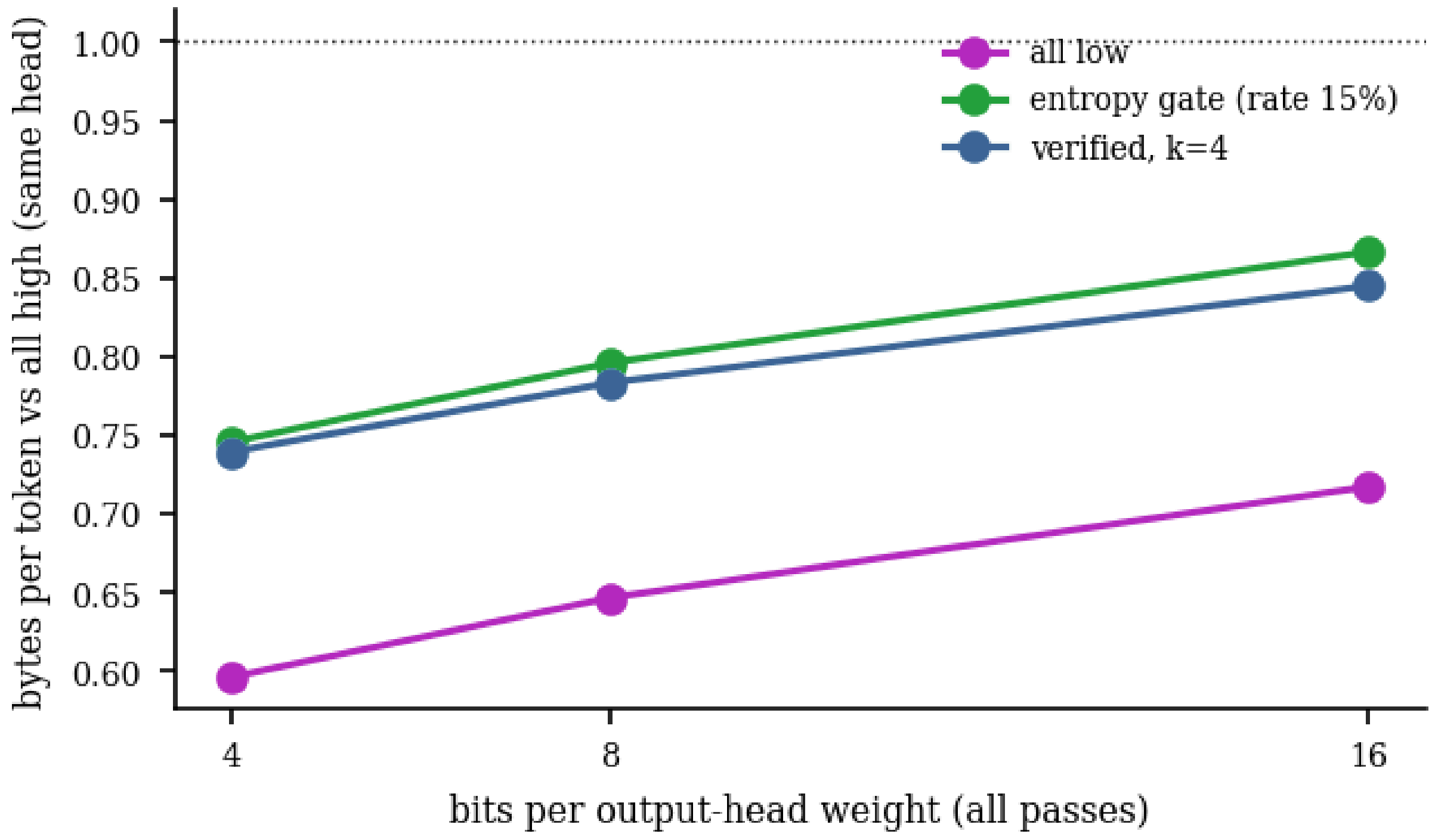

**Figure 20.** Relative bytes per token when the head is stored at 16, 8 or 4 bits in every pass, with the all-high baseline using the same head.

These costs cover weight reads only. Activations, the KV cache and kernel launch costs are not modelled, and wall-clock time was not used because the simulation has no packed kernels and the GPU was shared with other jobs. A real implementation would pay for dequantizing the low planes and might not reach the modelled saving.

## 9 Phase gating

Thinking-mode generations on Qwen3 contain a long reasoning span and a short answer. If the span tolerates low precision and the answer does not, a phase rule would spend high precision on a few tokens only, and it needs no uncertainty estimate, just the token that closes the span.

The run that would test this did not take place, so there is no evidence either way: the non-thinking generations used here contain no closing tag, and the replay analysis in Section 5.3 gives only a weak proxy through position effects. Three results would be needed before any claim: all-high, all-low and the phase rule on thinking-mode generations; the reverse phase rule; and a matched random control with the same share of high-precision tokens. The driver contains these modes and the cost model prices them, so the test can be run directly.

## 10 Statistical power of the live sample

The live comparison has 150 paired problems. Among them, 26 were right with all-high and wrong with all-low, and 13 went the other way, so the discordant share is 0.26. For a paired difference the approximate half-width of the 95% interval is 1.96 times the square root of the discordant share divided by the number of problems, which at this sample size is about 8.2 points. Figure 21 shows how it falls with n.

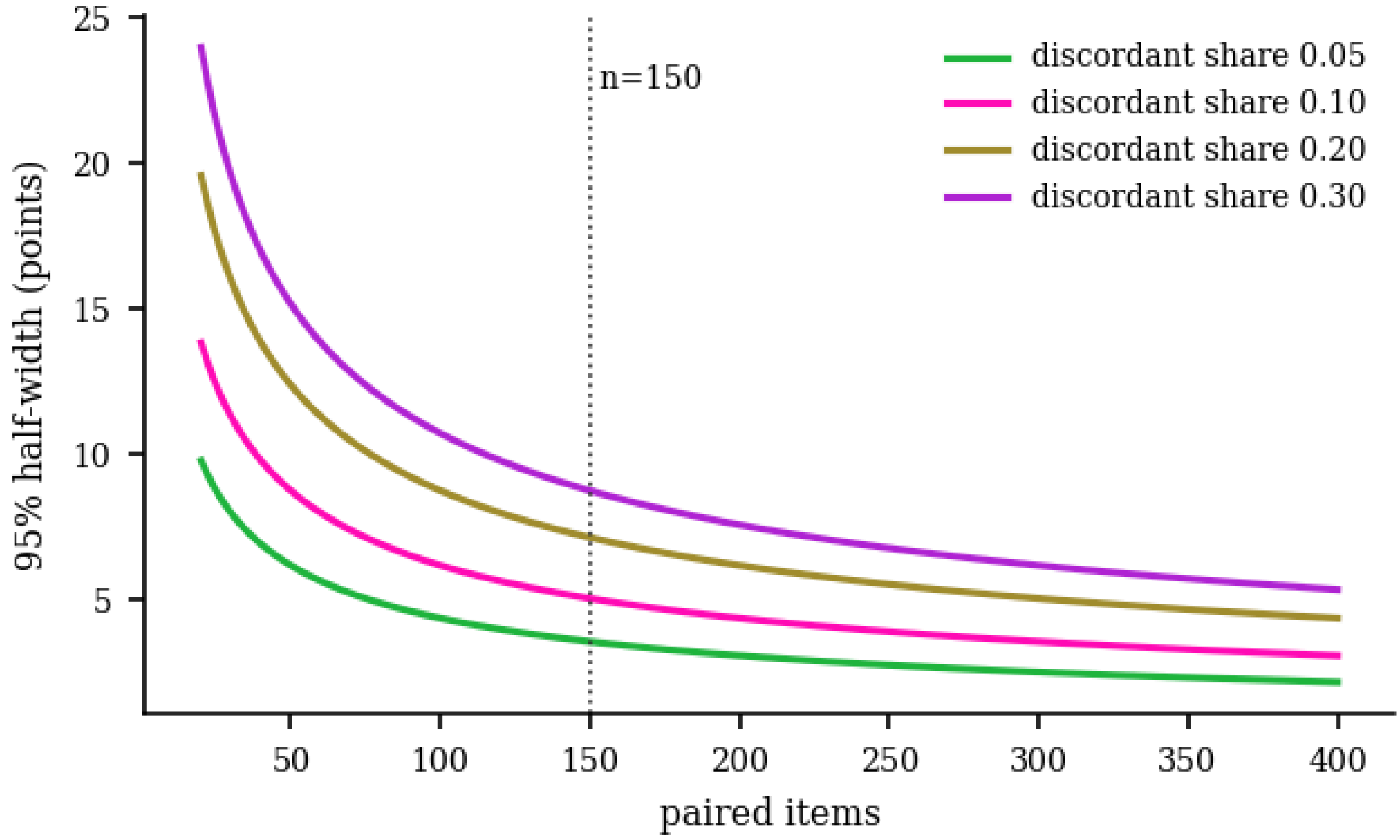

**Figure 21.** Approximate 95% half-width of a paired accuracy difference against the number of items, for several discordant shares. The dotted line marks the sample of this paper.

Take a gate that recovers half of the observed gap of 8.7 points between all-high and all-low. Telling it apart from all-low needs a half-width below 4.3 points, which at a discordant share of 0.26 means about 532 problems. The sample here is smaller, so confirmatory claims about accuracy need a larger run. The observed gap is itself an estimate with the interval shown in Table 7, so this figure is a rough guide and not a planned power calculation. The replay results rest on 36724 steps and are tighter, but they measure agreement of single steps and give no task accuracy.

Seeds do not enter decoding, since generation is greedy; randomness appears in the problem sample, in the bootstrap, and in the choice of 32 calibration problems. No second calibration set was tried, so the sensitivity of the thresholds to that choice is unknown.

## 11 Limitations

The weights are simulated round-to-nearest quantization with no calibration data, no outlier handling and no packed kernels. A tuned 4-bit method such as GPTQ or AWQ would have a smaller gap between low and high, and so less room for any gate. Trained nested weights [10] would behave differently again.

Only one small model was used, Qwen3-0.6B, and the 1.7B model that was planned did not run. Larger models, other families and mixture-of-experts models may have different entropy and flip structure. The benchmark is also single: GSM8K in non-thinking mode with a cap of 384 tokens and greedy decoding, where truncation affects accuracy and interacts with generation length.

The live run has 150 problems, so the intervals are wide and the tests are uncorrected across ten configurations. There is no live random control, no phase run and no live KL: random escalation is controlled in replay only, as an analytic expectation and not a sampled run, and the replay assumes that an escalated step equals the high step. Flips are counted per token, and the first departure of the all-low run has a median index of 2 tokens, so the flip counts do not say how many flips change an answer.

The cost model counts weight bytes only, with no wall-clock time, activations, KV traffic or dequantization cost. Thresholds come from 32 training problems and one setting of group size (64) and low bits (4). The related-work section rests on titles and abstracts of papers found by a limited search.

## 12 Discussion and next steps

For a reader who runs a small model locally and holds one 8-bit copy of the weights, the replay data say that a gate on the low-pass distribution finds the steps where the two precisions disagree far more often than chance would. At a 15% budget the entropy ordering reaches 78% of flips and the margin ordering 94%, and the oracle reaches 84%, so there is room above the simple gates.

The live accuracy data are too few to rank the gates reliably against each other, and the intervals above show how far each estimate can move. In the byte model, a gate that escalates 16% of steps costs 0.874 of the all-high bytes in the live run, a saving of 13% before any implementation cost. The accuracy price of that saving is -1.3 (-8.0 to +5.3) points against all-high, an interval that cannot exclude a loss of several points, and all-low saves 28% for a loss that is itself borderline, so the live run does not show that the gate buys accuracy that all-low lacks. Verified speculation with k=4 costs 0.845 in the same model and has zero KL, so a reader who can run verification should compare the two directly on their own hardware.

The simplest next steps are the matched random control and a thinking-mode run with the phase rules, on the same problems and the same cost model. A tuned low-bit quantizer and a lower-precision head would change both the gap to close and the price of closing it.

## Appendix A Reproduction and full table

Files: the decoder with simulated nested quantization, the driver, the analysis helpers and the paper-side analysis live in the local-model directory of the project (`a13_dec.py`, `a13_run.py`, `a13_ana.py`, `a13_paper.py`). Results are cached as one pickle per stage with one record per problem and configuration, plus a calibration file of thresholds and a metadata file with the byte counts.

Commands as run: the calibration pass decodes 32 training problems with the all-high model (`a13_run.py 0.6B gsm calib`). The live run is `a13_run.py 0.6B gsm s1 --cap 384 --batch 12 --n 150`. Group size 64 and 4 low bits are the defaults. The driver is resumable and skips finished jobs.

Every number in the text and the tables is recomputed at build time from these files, restricted to problems that finished in all ten configurations (live tables) or that have an all-high run (replay). Rebuilding after more problems finish would change the numbers.

**Table 8.** Per-configuration summary on the common problems, with absolute bytes per generated token (MB).

| Config | n | Accuracy (%) | Tokens | Capped (%) | Block MB/token | Total MB/token |
|---|---|---|---|---|---|---|
| high | 150 | 64.0 (56.7 to 71.3) | 244.8 | 10.7 | 469.8 | 781.0 |
| low | 150 | 55.3 (47.3 to 63.3) | 257.7 | 11.3 | 248.7 | 559.9 |
| ent0 | 150 | 62.0 (54.0 to 69.3) | 240.9 | 9.3 | 265.5 | 587.2 |
| ent1 | 150 | 62.0 (54.7 to 70.0) | 246.4 | 9.3 | 283.9 | 617.8 |
| ent2 | 150 | 62.7 (55.3 to 70.7) | 245.1 | 10.0 | 322.9 | 682.8 |
| ent3 | 150 | 65.3 (58.0 to 72.7) | 239.9 | 9.3 | 394.6 | 802.1 |
| mar0 | 150 | 59.3 (52.0 to 67.3) | 240.1 | 6.0 | 263.1 | 583.1 |
| mar1 | 150 | 61.3 (53.3 to 68.7) | 245.1 | 5.3 | 278.3 | 608.6 |
| mar2 | 150 | 66.0 (58.7 to 73.3) | 240.4 | 7.3 | 319.8 | 677.7 |
| mar3 | 150 | 64.7 (57.3 to 72.7) | 240.4 | 10.7 | 388.3 | 791.7 |

## References

1. Lotfi et al. Quantized Reasoning Models Think They Need to Think Longer, but They Do Not. arXiv:2606.00206, 2026.
2. Yang et al. BitNest: Bit-Nested Speculative Decoding for Memory-Efficient LLM Inference Acceleration. arXiv:2610.02800, 2026.
3. Xu et al. QuEPT: Quantized Elastic Precision Transformers with One-Shot Calibration for Multi-Bit Switching. arXiv:2602.12609, 2026.
4. Zhang et al. Speculative Decoding Meets Quantization: Compatibility Evaluation and Hierarchical Framework Design. arXiv:2505.22179, 2025.
5. Kim et al. Quantize the Target, Quantize the Drafter: Efficient Inference with Qwen3.5-4B. arXiv:2607.04244, 2026.
6. Song et al. MARS: Unleashing the Power of Speculative Decoding via Margin-Aware Verification. arXiv:2601.15498, 2026.
7. Haeri Boroujeni et al. Don’t Waste Bits! Adaptive KV-Cache Quantization for Lightweight On-Device LLMs. arXiv:2604.04722, 2026.
8. Park et al. Any-Precision LLM: Low-Cost Deployment of Multiple, Different-Sized LLMs. arXiv:2402.10517, 2024.
9. Simonds et al. Entropy Adaptive Decoding: Dynamic Model Switching for Efficient Inference. arXiv:2502.06833, 2025.
10. Nair et al. Matryoshka Quantization. arXiv:2502.06786, 2025.
11. Kleinegger et al. MatGPTQ: Accurate and Efficient Post-Training Matryoshka Quantization. arXiv:2602.03537, 2026.
12. Tiwari et al. QuantSpec: Self-Speculative Decoding with Hierarchical Quantized KV Cache. arXiv:2502.10424, 2025.
13. Huang et al. Quasar: Quantized Self-Speculative Acceleration for Rapid Inference via Memory-Efficient Verification. arXiv:2603.01399, 2026.
14. Jiang et al. TriSpec: Ternary Speculative Decoding via Lightweight Proxy Verification. arXiv:2601.23180, 2026.
15. Narasimhan et al. Faster Cascades via Speculative Decoding. arXiv:2405.19261, 2024.

16. Liu et al. Quantization Hurts Reasoning? An Empirical Study on Quantized Reasoning Models. arXiv:2504.04823, 2025.
17. Schuster et al. Confident Adaptive Language Modeling. arXiv:2207.07061, 2022.
18. Su et al. Entropy-Aware Speculative Decoding Toward Improved LLM Reasoning. arXiv:2512.23765, 2025.
19. Cobbe et al. Training Verifiers to Solve Math Word Problems. arXiv:2110.14168, 2021.
20. Yang et al. Qwen3 Technical Report. arXiv:2505.09388, 2025.
21. McNemar. Note on the sampling error of the difference between correlated proportions. *Psychometrika* 12, 1947.